\documentclass{article}

\PassOptionsToPackage{numbers, compress}{natbib}

\usepackage[final]{neurips_2023}

\usepackage[utf8]{inputenc} 
\usepackage[T1]{fontenc}    
\usepackage{hyperref}       
\usepackage{url}            
\usepackage{booktabs}       
\usepackage{amsfonts}       
\usepackage{nicefrac}       
\usepackage{microtype}      
\usepackage{xcolor}         

\usepackage{bm}
\usepackage{amsmath}
\usepackage{amssymb}
\usepackage{graphicx}
\usepackage{enumitem}
\usepackage{multirow}
\usepackage{colortbl}
\usepackage{booktabs}    
\usepackage{subfigure}
\usepackage{wrapfig}     
\usepackage{pdfpages}

\definecolor{mygray}{gray}{.85}
\DeclareMathOperator*{\argmin}{arg\,min}

\newcommand{\norm}[1]{\left\Vert#1\right\Vert}
\newcommand{\abs}[1]{\left\vert#1\right\vert}
\newcommand{\ang}[1]{\left\langle#1\right\rangle}
\newcommand{\etal}{\textit{et al}.}
\newcommand{\ie}{\textit{i}.\textit{e}.}
\newcommand{\eg}{\textit{e}.\textit{g}.}

\title{NeuralGF: Unsupervised Point Normal Estimation by Learning Neural Gradient Function}

\author{
  Qing Li$^1$ \quad
  Huifang Feng$^2$ \quad
  Kanle Shi$^3$ \quad
  Yue Gao$^1$ \quad
  Yi Fang$^4$ \\
  \textbf{Yu-Shen Liu}$^1$\thanks{Corresponding author} \quad
  \textbf{Zhizhong Han}$^5$  \\
  $^1$School of Software, Tsinghua University, Beijing, China \\
  $^2$School of Informatics, Xiamen University, Xiamen, China \\
  $^3$Kuaishou Technology, Beijing, China \\
  $^4$Center for Artificial Intelligence and Robotics, New York University Abu Dhabi, Abu Dhabi, UAE \\
  $^5$Department of Computer Science, Wayne State University, Detroit, USA \\
  \texttt{\{leoqli, gaoyue, liuyushen\}@tsinghua.edu.cn} \quad \texttt{fenghuifang@stu.xmu.edu.cn} \\
  \texttt{yfang@nyu.edu} \quad \texttt{h312h@wayne.edu}
}

\begin{document}

\maketitle

\begin{abstract}
  Normal estimation for 3D point clouds is a fundamental task in 3D geometry processing.
  The state-of-the-art methods rely on priors of fitting local surfaces learned from normal supervision.
  However, normal supervision in benchmarks comes from synthetic shapes and is usually not available from real scans, thereby limiting the learned priors of these methods.
  In addition, normal orientation consistency across shapes remains difficult to achieve without a separate post-processing procedure.
  To resolve these issues, we propose a novel method for estimating oriented normals directly from point clouds without using ground truth normals as supervision.
  We achieve this by introducing a new paradigm for learning neural gradient functions, which encourages the neural network to fit the input point clouds and yield unit-norm gradients at the points.
  Specifically, we introduce loss functions to facilitate query points to iteratively reach the moving targets and aggregate onto the approximated surface, thereby learning a global surface representation of the data.
  Meanwhile, we incorporate gradients into the surface approximation to measure the minimum signed deviation of queries, resulting in a consistent gradient field associated with the surface.
  These techniques lead to our deep unsupervised oriented normal estimator that is robust to noise, outliers and density variations.
  Our excellent results on widely used benchmarks demonstrate that our method can learn more accurate normals for both unoriented and oriented normal estimation tasks than the latest methods.
  The source code and pre-trained model are available at \textcolor{red}{\href{https://github.com/LeoQLi/NeuralGF}{https://github.com/LeoQLi/NeuralGF}}.
\end{abstract}

\section{Introduction}

Normal vectors are one of the local descriptors of point clouds and can offer additional geometric information for many downstream applications, such as denoising~\cite{sun2015denoising,lu2017gpf,lu2020low,lu2020deep}, segmentation~\cite{qi2017pointnet,qi2017pointnet++,qian2022pointnext} and registration~\cite{pomerleau2015review,serafin2015nicp}.
Moreover, some computer vision tasks require the normals to have consistent orientations, \ie, oriented normals,
such as graphics rendering~\cite{blinn1978simulation,gouraud1971continuous,phong1975illumination} and surface reconstruction~\cite{hoppe1992surface,kazhdan2005reconstruction,kazhdan2006poisson,kazhdan2013screened}.
Therefore, point cloud normal estimation has been an important research topic for a long time.
However, progress in this area has plateaued for some time, until the recent introduction of deep learning.
Currently, a growing body of work shows that performance improvements can be achieved using data-driven approaches~\cite{guerrero2018pcpnet,ben2020deepfit,li2022hsurf,li2023shsnet}, and these learning-based methods often give better results than traditional data-independent methods.
However, they suffer from shortcomings in terms of robustness to various data and computational efficiency~\cite{lenssen2020deep}.
Besides, they usually rely on a large amount of labeled training data and network parameters to learn high-dimensional features.
More importantly, the presence of varying noise levels, uneven sampling densities, and various complex geometries poses challenges for accurately estimating oriented normals from point clouds.

\begin{figure}
	\centering
	\includegraphics[width=\linewidth]{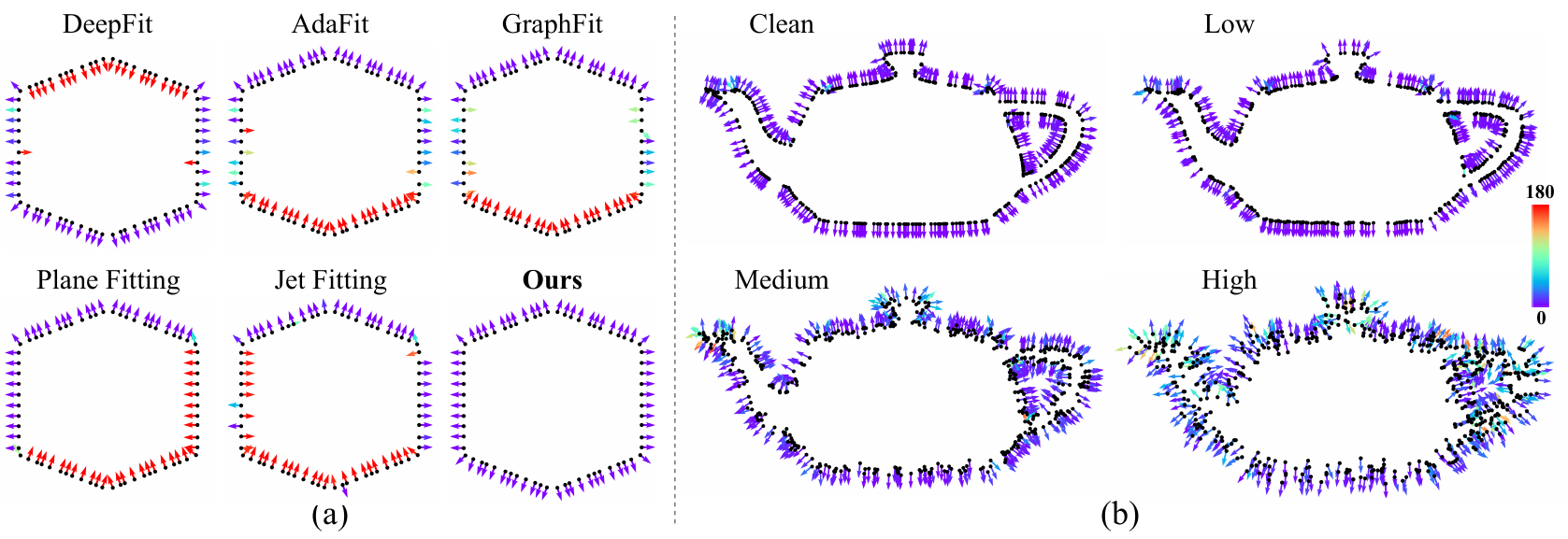}  \vspace{-0.7cm}
  \caption{
    (a) Existing surface fitting-based normal estimation methods can only produce unoriented normals (arrows) with inconsistent orientations.
    (b) Our method can estimate oriented normals (arrows) from unevenly sampled point clouds (black dots) with varying noise levels.
    The color of the arrow indicates the angle error compared to the ground truth.
  }
  \label{fig:fit_grad}
  \vspace{-0.2cm}
\end{figure}

Most existing point cloud normal estimation methods aim to locally fit geometric surfaces and compute normals from the fitted surfaces.
Recent surface fitting-based methods, such as DeepFit~\cite{ben2020deepfit}, AdaFit~\cite{zhu2021adafit} and GraphFit~\cite{li2022graphfit}, generalize the $n$-jet surface model~\cite{cazals2005estimating} to learning-based frameworks that predict pointwise weights using deep neural networks.
They obtain surface normals by solving the weighted least squares polynomial fitting problem.
However, as shown in Fig.~\ref{fig:fit_grad}(a), their methods can not guarantee consistent orientations very well, since $n$-jet only fits a local surface, which is ambiguous for determining orientation.
Furthermore, their methods need to construct local patches based on each query point for surface fitting.
This makes the algorithm need to traverse all patches in the point cloud, which is time-consuming.
More importantly, they rely on the priors learned from normal supervision to fit local surfaces, while such supervision is not always available.

Inspired by neural implicit representations~\cite{atzmon2020sal,atzmon2021sald,ma2020neural,ma2023towards,chen2023unsupervised}, we introduce {neural gradient functions} (NeuralGF) to approximate the global surface representation and constrain the gradient during the approximation.
Our method includes two basic principles:
(1) We formulate the neural gradient learning as iteratively learning an implicit surface representation directly from the entire point cloud for global surface fitting.
(2) We constrain the derived surface gradient vector to accurately describe the local distribution and make it consistent across iterations.
Accordingly, we implement these two principles using the designed loss functions, leading to a model that can provide accurate normals with consistent orientations from the entire point cloud (see Fig.~\ref{fig:fit_grad}(b)) through a single forward pass, and does not require ground truth labels or post-processing.
Our method improves the state-of-the-art results while using much fewer parameters.
To summarize, our main contributions include:
\begin{itemize}[leftmargin=*]
\setlength{\itemsep}{0pt}
\setlength{\parsep}{0pt}
\setlength{\parskip}{0pt}
\vspace{-0.2cm}
  \item Introducing a new paradigm of learning neural gradient functions, building on implicit geometric representations, for computing consistent gradients directly from raw data without ground truth.
  \item Implementing the paradigm with the losses designed to train neural networks to fit global surfaces and estimate oriented normals with consistent orientations in an end-to-end manner.
  \item Reporting the state-of-the-art performance for both unoriented and oriented normal estimation on point clouds with noise, density variations, and complex geometries.
\end{itemize}

\section{Related Work}

\noindent\textbf{Unoriented Normal Estimation}.
Point cloud normal estimation has been extensively studied over the past decades and widely used in many downstream applications.
In general, most of these studies focus on estimating unoriented normals, \ie, finding more accurate perpendiculars of local surfaces.
The Principle Component Analysis (PCA)~\cite{hoppe1992surface} is the most commonly used algorithm that has been widely applied to various geometric processing tasks.
Later, PCA variants~\cite{alexa2001point,pauly2002efficient,mitra2003estimating,lange2005anisotropic,huang2009consolidation}, Voronoi diagrams~\cite{amenta1999surface,dey2006provable,alliez2007voronoi,merigot2010voronoi} and Hough transform~\cite{boulch2012fast} have been proposed to improve the performance by preserving the shape details or sharp features.
Moreover, some methods~\cite{levin1998approximation,cazals2005estimating,guennebaud2007algebraic,oztireli2009feature,aroudj2017visibility} are designed to fit complex surfaces by containing more neighboring points, such as moving least squares~\cite{levin1998approximation}, truncated Taylor expansion ($n$-jet)~\cite{cazals2005estimating} and spherical surface fitting~\cite{guennebaud2007algebraic}.
These methods usually suffer from the parameters tuning, noise and outliers.
To achieve more robust capability, they usually choose a larger neighborhood size but over-smooth the details.
In recent years, deep learning-based normal estimation methods have been proposed and achieve great success.
The regression-based methods predict normals from structured data or raw point clouds.
Specifically, some methods~\cite{boulch2016deep,roveri2018pointpronets,lu2020deep} transform the 3D points into structured data, such as 2D grid representations.
Inspired by PointNet~\cite{qi2017pointnet}, other methods~\cite{guerrero2018pcpnet,zhou2020normal,hashimoto2019normal,ben2019nesti,zhou2020geometry,zhou2022refine,li2022hsurf,li2023NeAF,du2023rethinking,li2023shsnet} focus on the direct regression for unstructured point clouds.
In contrast to regressing the normal vectors directly, the surface fitting-based methods aim to train a neural network to predict pointwise weights, then the normals are solved through weighted plane fitting~\cite{lenssen2020deep,cao2021latent} or weighted polynomial surface fitting~\cite{ben2020deepfit,zhu2021adafit,zhou2023improvement,zhang2022geometry,li2022graphfit} on local neighborhoods.
Although these deep learning-based methods perform better than traditional data-independent methods, they require ground truth normals as supervision.
In this work, we propose to use networks to estimate normals in an unsupervised manner.

\noindent\textbf{Consistent Normal Orientation}.
Estimating normals with consistent orientations is also important and has been widely studied.
The most popular and simplest way for normal orientation is based on local propagation, where the orientation of a seed point is diffused to the adjacent points by a Minimum Spanning Tree (MST).
Most existing normal orientation approaches are based on this strategy, such as the pioneering work~\cite{hoppe1992surface} and its improved methods~\cite{konig2009consistent,seversky2011harmonic,wang2012variational,schertler2017towards,xu2018towards,jakob2019parallel}.
More recent work~\cite{metzer2021orienting} integrates a neural network to learn oriented normals within a single local patch and introduces a dipole propagation strategy across the partitioned patches to achieve global consistency.
These propagation-based approaches usually assume smoothness and suffer from local error propagation and the choice of neighborhood size.
Since local consistency is usually not enough to achieve robust orientations across different inputs,
some other methods determine the normal orientation by applying various volumetric representation techniques, such as signed distance functions~\cite{mello2003estimating,mullen2010signing}, variational formulations~\cite{walder2005implicit,alliez2007voronoi,huang2019variational}, visibility~\cite{katz2007direct,chen2010binary}, active contours~\cite{xie2004surface}, isovalue constraints~\cite{xiao2023point} and winding-number field~\cite{xu2023globally}.
In addition to the above approaches, the most recent researches~\cite{guerrero2018pcpnet,hashimoto2019normal,wang2022deep,li2023shsnet} aim to learn oriented normals directly from point clouds in a data-driven manner.
Similar to the unoriented normal estimation task, these deep learning-based methods perform better than traditional methods, but they require expensive ground truth as supervision.
In contrast, our method achieves better performance in an unsupervised manner.

\noindent\textbf{Neural Implicit Representation}.
Recent works~\cite{mescheder2019occupancy,park2019deepsdf} propose to learn neural implicit representation from point clouds.
Later, one class of methods~\cite{atzmon2020sal,atzmon2021sald,ma2020neural,peng2021shape,zhou2022learning,ma2022reconstructing,ma2022surface} focuses on training neural networks to overfit a single point cloud by introducing newly designed priors.
SAL~\cite{atzmon2020sal} uses unsigned regression to find signed local minima, thereby producing useful implicit representations directly from raw 3D data.
IGR~\cite{gropp2020implicit} proposes an implicit geometric regularization to learn smooth zero level set surfaces.
Neural-Pull~\cite{ma2020neural} uses signed distance to pull a point along the gradient to search for the shortest path to the underlying surface.
SAP~\cite{peng2021shape} proposes a differentiable Poisson solver to represent the surface as an oriented point cloud at the zero level set of an indicator function.
These methods focus on accurately locating the position of the zero iso-surface of an implicit function to extract the shape surface.
However, they ignore the constraints on the gradient of the function during optimizing the neural network.
We know that the gradient determines the direction of function convergence, and the gradient of the iso-surface can be used as the normal of the extracted surface.
If the gradient can be guided reasonably, the convergence process can be more robust and efficient, avoiding local extremum caused by noise or outliers.
Motivated by this, we try to incorporate neural gradients into implicit function learning to achieve accurate oriented normals.

\section{Method}

\noindent\textbf{Preliminary}.
Neural implicit representations are widely used in representing 3D geometry and have achieved promising performance in surface reconstruction.
Benefiting from the adaptability and approximation capabilities of deep neural networks~\cite{atzmon2019controlling}, recent approaches~\cite{park2019deepsdf,gropp2020implicit,ma2020neural,ma2022surface} use them to learn signed distance fields as a mapping from 3D coordinates $\bm{x}$ to signed distances.
These approaches represent surfaces as zero level sets of implicit functions $f$, \ie,
$\mathcal{S} = \left \{\bm{x} \in \mathbb{R}^3 ~ \vert ~ f(\bm{x}; \bm{\theta}) = 0 \right \}$,
where $f\colon\mathbb{R}^{3} \!\rightarrow\! \mathbb{R}$ is a network with parameter $\bm{\theta}$, which is usually implemented as a multi-layer perceptron (MLP).
If the function $f$ is continuous and differentiable, the normal of a point $p$ on the surface can be represented as $\mathbf{n}_{p} \!=\! \nabla f(p)/\norm{\nabla f(p)}$, where $\norm{\cdot}$ denotes the euclidean $L^2$-norm and $\nabla f(p)$ indicates the gradient at $p$, which can be obtained in the differentiation process of $f$.
Specifically, there are some important properties of the gradient:
(1) The maximum rate of change of the function $f$ is defined by the magnitude of the gradient $\norm{\nabla f}$ and given by the direction of $\nabla f$.
(2) The gradient vector $\nabla f(p)$ is perpendicular to the level surface $f(p) \!=\! 0$.

\begin{figure}
	\centering
	\includegraphics[width=\linewidth]{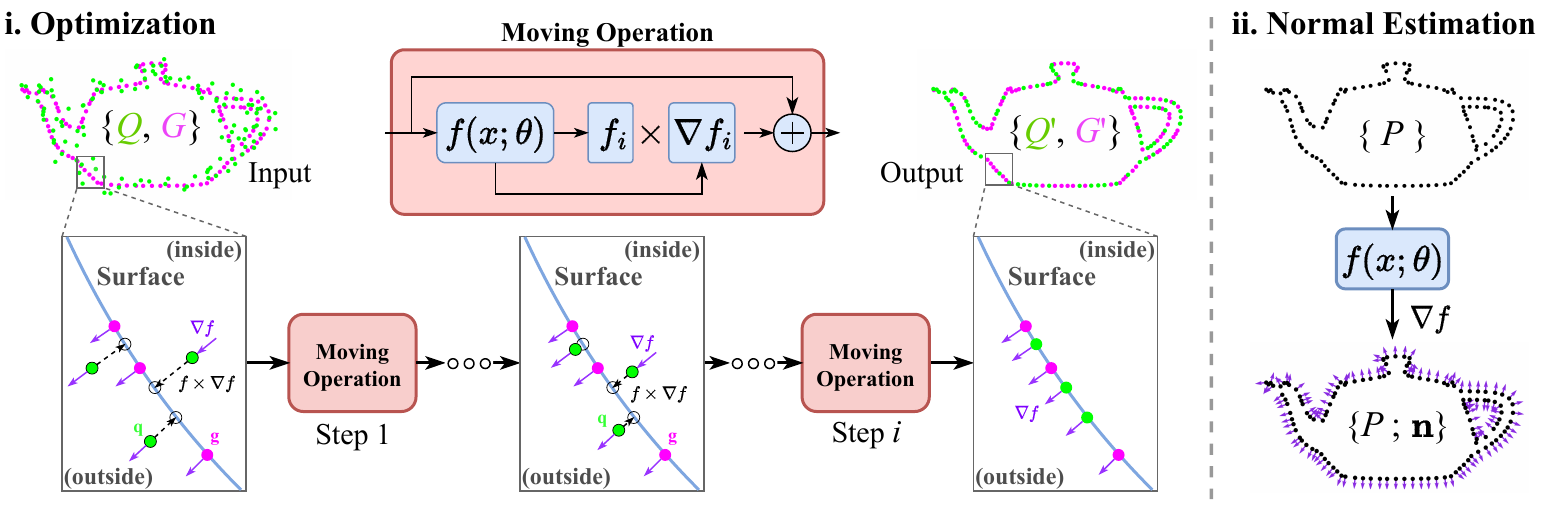}  \vspace{-0.6cm}
  \caption{
    Overview of our method.
    The optimization is formulated as an iterative moving operation of the positions of point set $\{Q,G\}$, which is sampled from the raw data $P$, while the normal estimation is a single forward differentiation of the learned function $f(x;\theta)$ using $P$ as input.
    We move the input set $\{Q,G\}$ through multiple steps using signed distance $f_i$ and gradient $\nabla f_i$ learned from $f(x;\theta)$ to obtain the output set $\{Q',G'\}$.
    The gradient is initialized to point outside the surface.
  }
  \label{fig:net}
  \vspace{-0.3cm}
\end{figure}

\noindent\textbf{Overview}.
We aim to train a neural network $f$ with parameter $\bm{\theta}$ from input points $\bm{x}$ to obtain the gradient $\nabla f$ during the optimization of $f$.
The optimization encourages $f$ to fit $\bm{x}$ and the gradient $\nabla f$ to approach the normal of the implicit surface.
We sample a query point set $Q \!=\! \{q_j\}_{j=1}^{N_Q}$ and a surface point set $G \!=\! \{g_j\}_{j=1}^{N_G}$ from raw point cloud $P$.
Specifically, $Q$ is sampled through some probability distribution $\mathcal{D}$ from $P$~\cite{cai2020learning,ma2020neural}.
To explicitly indicate the underlying geometry, we sample the point set $G \!\subset\! P$ that may be distorted by noise, and we verify that it can facilitate the learning of $f$ (see Sec.~\ref{sec:ablation}).
As shown in Fig.~\ref{fig:net}, our optimization pipeline is formulated as an iterative moving operation of the union of points $Q$ and $G$, more specifically, with $\mathcal{I}$ steps of position movement.
During normal estimation, the entire point cloud $P$ is fed into the network to derive the gradient embedded in the learned function.
To learn the neural gradient function based on the implicit surface representation of $f$, we consider a loss of the form
\begin{equation}
  \mathcal{L}(\bm{\theta}) = \mathbb{E}_{\bm{x} \sim \mathcal{D}}
  \Big( \mathcal{T} \big(V(\bm{x}; \bm{\theta}), ~\mathcal{V}(\bm{x}) \big)
        + \lambda ~\mathcal{T} \big(F(\bm{x}; \bm{\theta}), ~\mathcal{F}(\bm{x}) \big) \Big),
\end{equation}
where $\mathcal{T}\colon\mathbb{R}\times \mathbb{R} \rightarrow \mathbb{R}$ is a differentiable similarity function, and we adopt the standard $L^2$ (Euclidean) distance.
$V(\bm{x}; \bm{\theta})$ and $F(\bm{x}; \bm{\theta})$ represent the approximated gradient vector and global surface representation of the underlying geometry $P$, respectively.
$\mathcal{V}(\bm{x})$ and $\mathcal{F}(\bm{x})$ denote the corresponding measures, respectively.
We will introduce them in the following two sections.
$\lambda>0$ is a weighting factor.
The expectation $\mathbb{E}$ is made for certain probability distribution $\bm{x}\sim\mathcal{D}$ in 3D space.

\subsection{Learning Global Surface}
The $n$-jet surface model represents a polynomial function that maps points $(x,y) \!\in\! \mathbb{R}^2$ to their height $z$ that is not in the tangent space over a local surface~\cite{ben2020deepfit,zhu2021adafit,li2022graphfit}.
In contrast, we aim to learn an implicit global surface representation of the point cloud.
The gradient indicates the direction in 3D space in which the signed distance from the surface increases the fastest, so moving a point along the gradient will find its nearest path to the surface.
According to this property, we adopt a moving operation introduced in~\cite{ma2020neural} to project a query point $q$ to $q'$, and $q'= q - f(q; \bm{\theta}) \cdot \mathbf{n}_q$, $\mathbf{n}_q \!=\! \nabla f(q; \bm{\theta}) / \norm{\nabla f(q; \bm{\theta})}$.
We also use $\nabla f$ to denote the normalized gradient for simplification.
For the point $q\in Q$, we expect the function $f$ could provide correct signed distance $f(q; \bm{\theta})$ and gradient $\nabla f(q; \bm{\theta})$ that can be used to move $q$ to the nearest point $\hat{q}$ over the surface, and we minimize the error $\norm{q' - \hat{q}}$.
In this way, the function $f$ can be used as an implicit representation of the underlying surface and the zero level set of $f$ will be a valid manifold describing the point cloud.
In this work, the moving operation for input point set $\{Q, G\}$ is formulated as
{\small
\begin{equation} \label{eq:point_moving}
  \mathcal{L}_{d}(\bm{\theta}) = \mathbb{E}_{\bm{x} \sim \mathcal{D}}
  \mathcal{T} \big(F(\bm{x}; \bm{\theta}), \mathcal{F}(\bm{x}) \big) = \norm{\{Q',G'\} - \{\hat{Q},\hat{G}\}},
  ~~ \{Q',G'\} = \{Q,G\} - \sum_{i=1}^{\mathcal{I}} f_i \cdot \nabla f_i ,
\end{equation}
}
where $\{Q',G'\}$ is the position of point set $\{Q,G\}$ after moving in multiple steps $\mathcal{I}$ (see Fig.~\ref{fig:net}), and $\{\hat{Q},\hat{G}\}$ is the target position.
$f_i$ represents the signed distance of points during the $i$-th position movement step.
In practice, we find the nearest point of $Q$ and $G$ from the raw point cloud $P$ as the target position.
Different from previous works~\cite{ma2020neural,ma2022reconstructing,ma2022surface}, we introduce a multi-step movement strategy to optimize the point set $\{Q', G'\}$ to cover the surface instead of single-step movement.
The insight of our design is that it is always difficult to directly reach the target in one step if some points are distributed far from the surface or distorted by noise.
At this time, the optimization of the moving operation may go through many twists and turns to converge.

We optimize $\mathcal{L}_{d}(\bm{\theta})$ in an end-to-end manner based on the moved point set $\{Q',G'\}$.
For the points $G \subset P$ in the first step, we expect them to be on the underlying surface.
For the points $\{Q',G'\}$ moved after the first step, we expect them to be as close to the underlying surface as possible.
Hence, we leverage a regression loss term on the predicted distances $f_i^Q$ and $f_i^G$ of point set $\{Q,G\}$, \ie,
\begin{equation}
  \mathcal{L}_{reg}(\bm{\theta}) = \big( f_1^G \big)^2 + \sum_{i=2}^{\mathcal{I}} ( f_i )^2,
  ~~ {\rm and} ~ f_i = \big\{ f_i^Q, f_i^G \big\}.
\end{equation}

\subsection{Learning Consistent Gradient}
A standard way for estimating unoriented point normals is to fit a plane to the local neighborhood $\mathcal{N}(p, P)$ of each point $p$~\cite{levin1998approximation}.
Given a neighborhood size $K$, we let $\mathcal{V}(p) \in \mathbb{R}^{K \times 3}$ denote the set of centered coordinates of points in that neighborhood.
Then the plane fitting is described as finding the least squares solution of
\begin{equation} \label{eq:plane}
  \mathbf{n}_p^\ast = \argmin_{\mathbf{n}:\norm{\mathbf{n}}=1} \sum_{k=1}^{K} \norm{\mathbf{n} \cdot \mathcal{V}(p)_k}^2,
  ~~~{\rm and}~~ \mathcal{V}(p)_k = p_k - \frac{1}{K} \sum_{l=1}^{K} p_l ~,~~~p_k, p_l \in \mathcal{N}(p, P).
\end{equation}
However, the inner product in the formula allows the distribution of normals on both sides of the surface (positive or negative orientations) to satisfy the optimization.
Therefore, this approach cannot determine the normal orientation with respect to the surface, and a consistent orientation needs to be solved by post-processing, such as MST-based orientation propagation.
In this work, we directly estimate oriented normals by incorporating neural gradients into implicit function learning~\cite{atzmon2020sal,atzmon2021sald},
\begin{equation} \label{eq:grad}
  \mathcal{L}_{v}(\bm{\theta}) = \mathbb{E}_{\bm{x} \sim \mathcal{D}}
  \mathcal{T} \big(V(\bm{x}; \bm{\theta}), ~\mathcal{V}(\bm{x}) \big)
  = \norm{f(\bm{x}; \bm{\theta}) \cdot \mathbf{n} - \mathcal{V}(\bm{x})},
\end{equation}
where $\mathbf{n}$ is the unit gradient (normal) of each point on the surface.
The above formula aims to measure the minimum deviation of each point from the underlying surface.
Specifically, the gradient is perpendicular to the surface and the distance from the surface increases the fastest along the direction of the gradient.
Therefore, we measure the deviation from $P$ to $Q$ by finding the optimal gradients.
Generally, raw point clouds may contain noise and outliers that can severely degrade the accuracy of the surface approximation.
To reduce the impact of inaccuracy in $P$, we introduce multi-scale neighborhood size $\{K_s\}^{N_K}_{s=1}$ for $\mathcal{V}(x)$ based on a statistical manner, that is
\begin{equation} \label{eq:scale}
  \mathcal{V}_s(q) = q - \frac{1}{K_s}\sum_{k=1}^{K_s} q_k, ~~q_k \in \mathcal{N}_{K_s}(q, P), ~~ s=1,\cdots,N_K,
\end{equation}
where $\mathcal{N}_{K_s}(q, P)$ denotes the $K_s$ nearest points of $q \in Q$ in $P$.
$\mathcal{V}(q)$ is built as a vector from the averaged position $\bar{\bm{x}} \!=\! \sum_{k=1}^{K_s} \mathcal{N}_{K_s}(q, P) / K_s$ on surface to query $q$.
Then, the objective of Eq.~\eqref{eq:grad} is turned into the aggregation of the error at each neighborhood size scale, \ie,
\begin{equation} \label{eq:loss_grad}
  \mathcal{L}_{v}(\bm{\theta}) = \mathbb{E}_{\bm{x} \sim \mathcal{D}}
  \mathcal{T} \big(V(\bm{x}; \bm{\theta}), ~\mathcal{V}(\bm{x}) \big)
  = \sum_{s=1}^{N_K} \norm{f_i^Q \cdot \nabla f_i^Q - \mathcal{V}_s(Q)},
\end{equation}
where $i \!=\! 1,\cdots,\mathcal{I}$ is the number of steps.
In practice, we only use $f_1$ of the first step and its $\nabla f_1$ to measure the deviation, since we want the optimization to converge as fast as possible, which also implies that the position $Q'$ after the first move step in Eq.~\eqref{eq:point_moving} is as close to the surface as possible.

In addition, we pursue better gradient consistency across iterations, \ie, make the gradient directions in each move step parallel.
To achieve this, we align the gradient $\nabla f_i$ after the first step with the gradient $\nabla f_1$ of the first step.
Meanwhile, we also consider the confidence of each point with respect to the underlying surface.
Thus, we introduce confidence-weighted cosine distance to evaluate the gradient consistency at points $\{Q,G\}$, with the form of
\begin{equation} \label{eq:loss_con}
  \mathcal{L}_{con}(\bm{\theta}) = \sum_{i=2}^{\mathcal{I}} w \cdot \big(1 - \ang{\nabla f_1, \nabla f_i} \big),
  ~~ {\rm and} ~ \nabla f_i = \big\{\nabla f_i^Q, \nabla f_i^G \big\},
\end{equation}
where $\ang{\cdot}$ denotes cosine distance of vectors.
$w \!=\! {\rm exp}(-\rho \cdot \abs{f_1})$ is an adaptive weight that indicates the significance of each input point based on the predicted distance, making the model pay more attention to points near the surface.
In summary, our method combines an implicit representation and a gradient approximation with respect to the underlying geometry of $P$.
We not only consider the gradient consistency at $q$ during iterations, but also unify the adjacent regions of $q$ to compensate for the inaccuracy of $P$.
This is important because the input point cloud may be noisy and some points are not on the underlying surface.

\noindent\textbf{Loss Function}.
We optimize global surface approximation and consistent gradient learning in an end-to-end manner by adjusting the parameter $\bm{\theta}$ of $f$ using the following loss function
\begin{equation} \label{eq:loss}
  \mathcal{L}(\bm{\theta}) = \mathcal{L}_{v}(\bm{\theta}) + \lambda_1 \mathcal{L}_{con}(\bm{\theta}) + \lambda_2 \mathcal{L}_{d}(\bm{\theta}) + \lambda_3 \mathcal{L}_{reg}(\bm{\theta}),
\end{equation}
where $\lambda_1=0.01$, $\lambda_2=0.1$ and $\lambda_3=10$ are balance weights and are fixed across all experiments.

\begin{figure}[t]
	\centering
	\includegraphics[width=\linewidth]{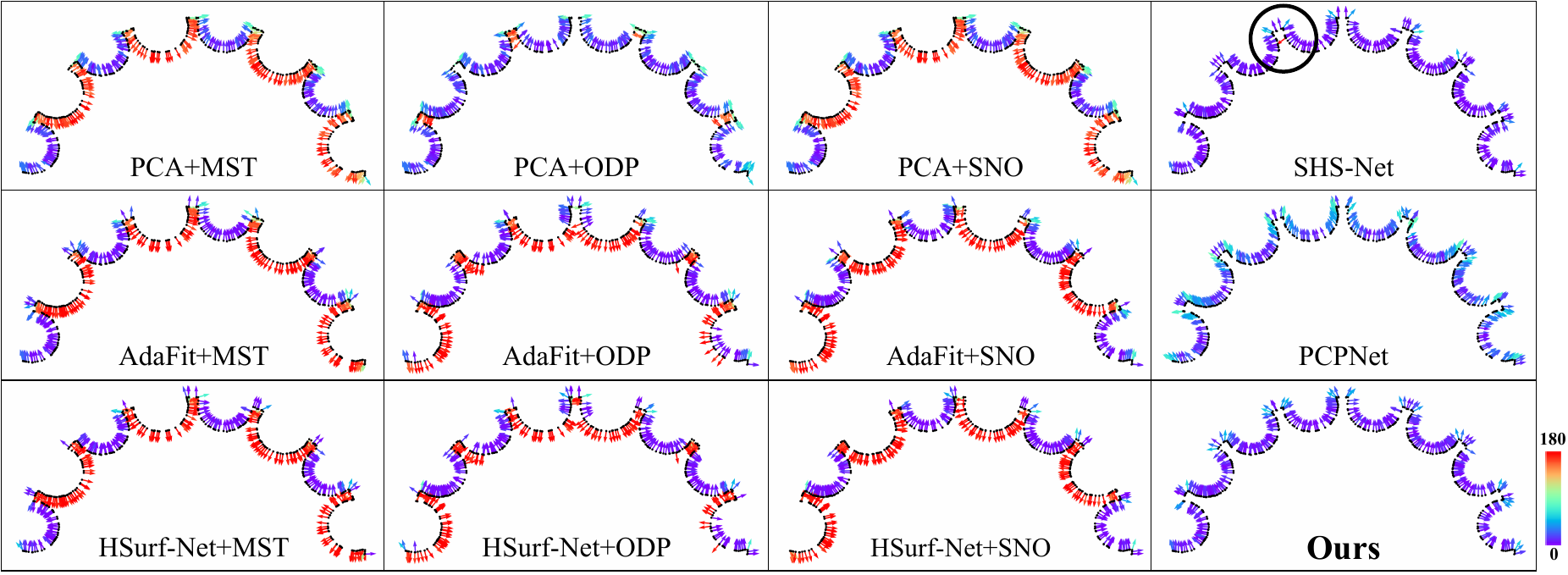}  \vspace{-0.6cm}
  \caption{
    Visual comparison of oriented normals.
    We only show a partial cross-section of a closed shape for better visualization.
    The color of the arrow indicates RMSE, where purple means the same orientation as the ground truth and red means the opposite.
  }
  \label{fig:error_gradient}
  \vspace{-0.2cm}
\end{figure}

\begin{table*}[t]
\centering
\footnotesize
\setlength{\tabcolsep}{1.0mm}
\caption{
	RMSE of oriented normals on datasets PCPNet and FamousShape.
	Our method achieves SOTA performance even when compared to supervised baselines.
	$\ast$ means the code is uncompleted.
}
\label{table:pcpnet_famousShape_o}
\resizebox{\linewidth}{!}{
\begin{tabular}{l|cccc|cc| >{\columncolor{mygray}} c||  cccc|cc| >{\columncolor{mygray}} c}
	\toprule
	\multirow{3}{*}{Category} & \multicolumn{7}{c||}{\textbf{PCPNet Dataset}} & \multicolumn{7}{c}{\textbf{FamousShape Dataset}} \\
	\cmidrule(r){2-15}
	& \multicolumn{4}{c|}{Noise} & \multicolumn{2}{c|}{Density} &     & \multicolumn{4}{c|}{Noise} & \multicolumn{2}{c|}{Density} &    \\
	& None & 0.12\% & 0.6\% & 1.2\%  & Stripe & Gradient  & \multirow{-2}{*}{{Average}} & None & 0.12\% & 0.6\% & 1.2\%  & Stripe & Gradient & \multirow{-2}{*}{{Average}} \\
	\midrule
	PCA~\cite{hoppe1992surface}+MST~\cite{hoppe1992surface}
	& 19.05 & 30.20 & 31.76 & 39.64 & 27.11 & 23.38 &   28.52    & 35.88 & 41.67 & 38.09 & 60.16 & 31.69 & 35.40 &   40.48  \\
	PCA~\cite{hoppe1992surface}+SNO~\cite{schertler2017towards}
	& 18.55 & 21.61 & 30.94 & 39.54 & 23.00 & 25.46 &   26.52    & 32.25 & 39.39 & 41.80 & 61.91 & 36.69 & 35.82 &   41.31  \\
	PCA~\cite{hoppe1992surface}+ODP~\cite{metzer2021orienting}
	& 28.96 & 25.86 & 34.91 & 51.52 & 28.70 & 23.00 &   32.16    & 30.47 & 31.29 & 41.65 & 84.00 & 39.41 & 30.72 &   42.92  \\
	LRR~\cite{zhang2013point}+MST~\cite{hoppe1992surface}
	& 43.48 & 47.58 & 38.58 & 44.08 & 48.45 & 46.77 &   44.82    & 56.24 & 57.38 & 45.73 & 64.63 & 66.35 & 56.65 &   57.83  \\
	LRR~\cite{zhang2013point}+SNO~\cite{schertler2017towards}
	& 44.87 & 43.45 & 33.46 & 45.40 & 46.96 & 37.73 &   41.98    & 59.78 & 60.18 & 45.02 & 71.37 & 62.78 & 59.90 &   59.84  \\
	LRR~\cite{zhang2013point}+ODP~\cite{metzer2021orienting}
	& 28.65 & 25.83 & 36.11 & 53.89 & 26.41 & 23.72 &   32.44    & 39.97 & 42.17 & 48.29 & 88.68 & 44.92 & 47.56 &   51.93  \\
	AdaFit~\cite{zhu2021adafit}+MST~\cite{hoppe1992surface}
	& 27.67 & 43.69 & 48.83 & 54.39 & 36.18 & 40.46 &   41.87    & 43.12 & 39.33 & 62.28 & 60.27 & 45.57 & 42.00 &   48.76  \\
	AdaFit~\cite{zhu2021adafit}+SNO~\cite{schertler2017towards}
	& 26.41 & 24.17 & 40.31 & 48.76 & 27.74 & 31.56 &   33.16    & 27.55 & 37.60 & 69.56 & 62.77 & 27.86 & 29.19 &   42.42  \\
	AdaFit~\cite{zhu2021adafit}+ODP~\cite{metzer2021orienting}
	& 26.37 & 24.86 & 35.44 & 51.88 & 26.45 & 20.57 &   30.93    & 41.75 & 39.19 & 44.31 & 72.91 & 45.09 & 42.37 &   47.60  \\
	HSurf-Net~\cite{li2022hsurf}+MST~\cite{hoppe1992surface}
	& 29.82 & 44.49 & 50.47 & 55.47 & 40.54 & 43.15 &   43.99    & 54.02 & 42.67 & 68.37 & 65.91 & 52.52 & 53.96 &   56.24  \\
	HSurf-Net~\cite{li2022hsurf}+SNO~\cite{schertler2017towards}
	& 30.34 & 32.34 & 44.08 & 51.71 & 33.46 & 40.49 &   38.74    & 41.62 & 41.06 & 67.41 & 62.04 & 45.59 & 43.83 &   50.26  \\
	HSurf-Net~\cite{li2022hsurf}+ODP~\cite{metzer2021orienting}
	& 26.91 & 24.85 & 35.87 & 51.75 & 26.91 & 20.16 &   31.07    & 43.77 & 43.74 & 46.91 & 72.70 & 45.09 & 43.98 &   49.37  \\
	PCPNet~\cite{guerrero2018pcpnet}
	& 33.34 & 34.22 & 40.54 & 44.46 & 37.95 & 35.44 &   37.66    & 40.51 & 41.09 & 46.67 & 54.36 & 40.54 & 44.26 &   44.57  \\
	DPGO$^\ast$~\cite{wang2022deep}
	& 23.79 & 25.19 & 35.66 & 43.89 & 28.99 & 29.33 &   31.14    & - & - & - & - & - & - & -  \\
	SHS-Net~\cite{li2023shsnet}
	& \textbf{10.28} & \textbf{13.23} & 25.40 & 35.51 & 16.40 & 17.92 &   19.79    & 21.63 & 25.96 & 41.14 & 52.67 & 26.39 & 28.97 &   32.79  \\
	Ours
	&        {10.60} &        {18.30} & \textbf{24.76} & \textbf{33.45} & \textbf{12.27} & \textbf{12.85} &   \textbf{18.70}
	& \textbf{16.57} & \textbf{19.28} & \textbf{36.22} & \textbf{50.27} & \textbf{17.23} & \textbf{17.38} &   \textbf{26.16}  \\
	\bottomrule
\end{tabular} }
\vspace{-0.4cm}
\end{table*}

\noindent\textbf{Neural Gradient Computation}.
The above losses require incorporating the gradient $\nabla f(\bm{x};\bm{\theta})$ in a differentiable manner.
As in~\cite{gropp2020implicit,atzmon2021sald}, we construct $\nabla f(\bm{x};\bm{\theta})$ as a neural network in conjunction with $f(\bm{x};\bm{\theta})$.
Specifically, each layer of the network has the form $\bm{y}^{\ell+1} \!=\! \phi(\bm{W} \bm{y}^{\ell} + \bm{b})$,
where $\bm{\theta} \!=\! (\bm{W}, \bm{b})$ is the learnable parameter of $\ell$ layer with output $\bm{y}^{\ell+1}$, and $\phi:\mathbb{R}\rightarrow\mathbb{R}$ is a non-linear derivable activation function.
According to chain-rule, the gradient is given by~\cite{gropp2020implicit}
\begin{equation} \label{eq:grads}
  \nabla \bm{y}^{\ell+1} = \textrm{diag} \big(\phi' (\bm{W} \bm{y}^{\ell} + \bm{b} ) \big) \bm{W} \nabla \bm{y}^{\ell},
\end{equation}
where $\phi'$ is the derivative of $\phi$, and $\textrm{diag}(\bm{z})$ arrange its input vector $\bm{z}\in\mathbb{R}^m$ on the diagonal of a square matrix $\mathbb{R}^{m\times m}$.
In practice, we use the automatic differentiation module of PyTorch~\cite{paszke2017automatic} to implement the computation of the gradient $\nabla f(\bm{x};\bm{\theta})$ in the forward pass.

\begin{figure}
	\centering
	\includegraphics[width=\linewidth]{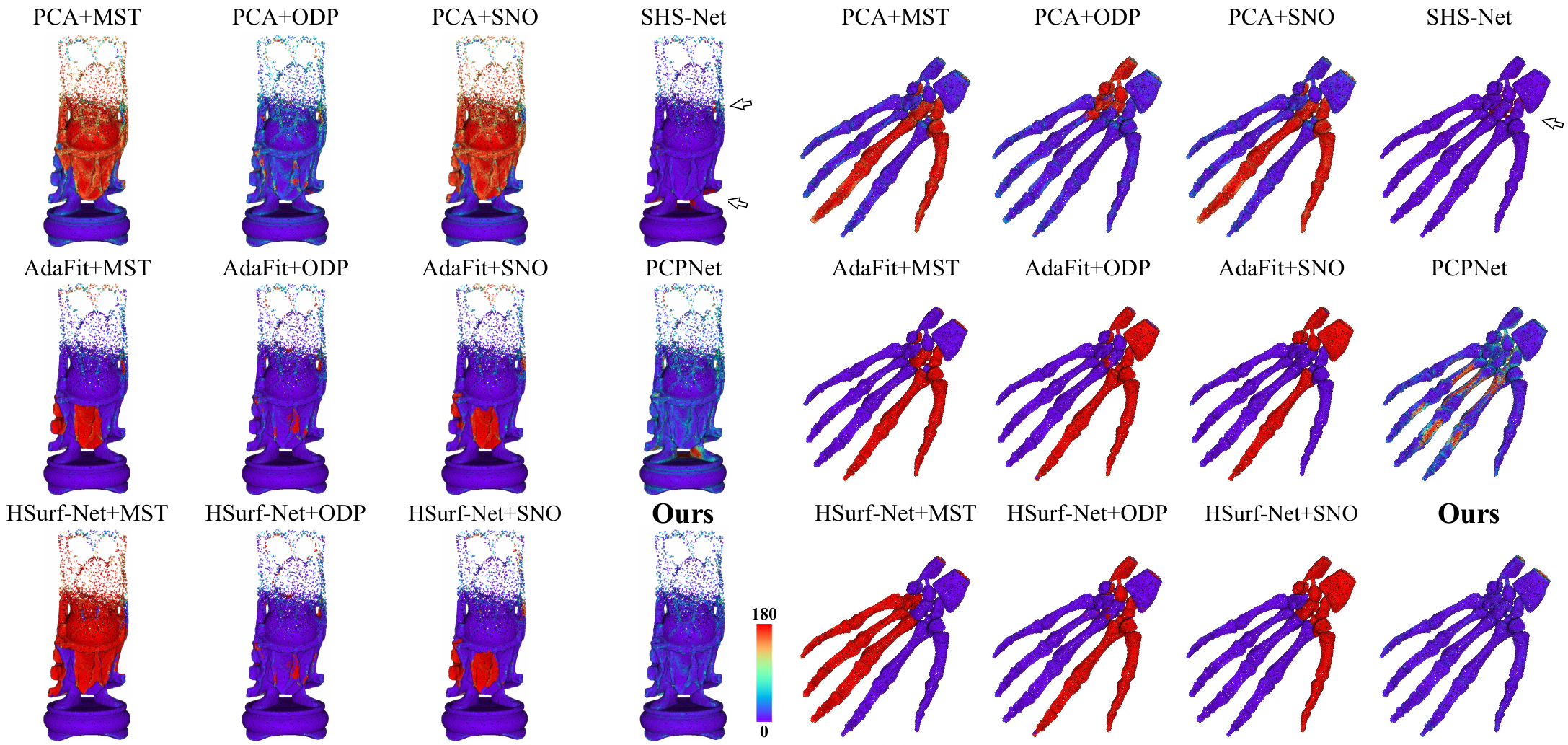}  \vspace{-0.6cm}
  \caption{
    Visual comparison of oriented normals on point clouds with density variations and complex geometries.
    Point color indicates normal RMSE and red ($180^\circ$) indicates reversed orientation.
  }
  \label{fig:error_map}
  \vspace{-0.2cm}
\end{figure}

\begin{table*}[t]
\centering
\footnotesize
\setlength{\tabcolsep}{1.0mm}
\caption{
	RMSE of unoriented normal on datasets PCPNet and FamousShape.
	Our method achieves SOTA performance compared to unsupervised baseline methods.
}
\label{table:pcpnet_famousShape}
\resizebox{\linewidth}{!}{
\begin{tabular}{l|cccc|cc| >{\columncolor{mygray}} c||  cccc|cc| >{\columncolor{mygray}} c}
	\toprule
	\multirow{3}{*}{Category} & \multicolumn{7}{c||}{\textbf{PCPNet Dataset}} & \multicolumn{7}{c}{\textbf{FamousShape Dataset}} \\
	\cmidrule(r){2-15}
	& \multicolumn{4}{c|}{Noise} & \multicolumn{2}{c|}{Density} &     & \multicolumn{4}{c|}{Noise} & \multicolumn{2}{c|}{Density} &    \\
	& None & 0.12\% & 0.6\% & 1.2\%  & Stripe & Gradient  & \multirow{-2}{*}{{Average}} & None & 0.12\% & 0.6\% & 1.2\%  & Stripe & Gradient & \multirow{-2}{*}{{Average}} \\
	\midrule
	\textbf{Supervised} &&&&&&&&&&&&&& \\
	PCPNet~\cite{guerrero2018pcpnet}             & 9.64  & 11.51 & 18.27 & 22.84 & 11.73 & 13.46 &  14.58    & 18.47 & 21.07 & 32.60 & 39.93 & 18.14 & 19.50 &   24.95  \\
	Nesti-Net~\cite{ben2019nesti}                & 7.06  & 10.24 & 17.77 & 22.31 & 8.64  & 8.95  &  12.49    & 11.60 & 16.80 & 31.61 & 39.22 & 12.33 & 11.77 &   20.55  \\
	Lenssen \etal~\cite{lenssen2020deep}         & 6.72  & 9.95  & 17.18 & 21.96 & 7.73  & 7.51  &  11.84    & 11.62 & 16.97 & 30.62 & 39.43 & 11.21 & 10.76 &   20.10  \\
	DeepFit~\cite{ben2020deepfit}                & 6.51  & 9.21  & 16.73 & 23.12 & 7.92  & 7.31  &  11.80    & 11.21 & 16.39 & 29.84 & 39.95 & 11.84 & 10.54 &   19.96  \\
	Zhang \etal~\cite{zhang2022geometry}         & 5.65  & 9.19  & 16.78 & 22.93 & 6.68  & 6.29  &  11.25    & 9.83 & 16.13 & 29.81 & 39.81 & 9.72 & 9.19 &   19.08  \\
	AdaFit~\cite{zhu2021adafit}                  & 5.19  & 9.05  & 16.45 & 21.94 & 6.01  & 5.90  &  10.76    & 9.09 & 15.78 & 29.78 & 38.74 & 8.52 & 8.57 &   18.41  \\
	GraphFit~\cite{li2022graphfit}               & 5.21  & 8.96  & 16.12 & 21.71 & 6.30  & 5.86  &  10.69    & 8.91 & 15.73 & 29.37 & 38.67 & 9.10 & 8.62 &   18.40  \\
	NeAF~\cite{li2023NeAF}                       & 4.20  & 9.25  & 16.35 & 21.74 & 4.89  & 4.88  &  10.22    & 7.67 & 15.67 & 29.75 & 38.76 & 7.22 & 7.47 &   17.76  \\
	HSurf-Net~\cite{li2022hsurf}                 & 4.17  & 8.78  & 16.25 & 21.61 & 4.98  & 4.86  &  10.11    & 7.59 & 15.64 & 29.43 & 38.54 & 7.63 & 7.40 &   17.70  \\
	Du \etal~\cite{du2023rethinking}             & 3.85  & 8.67  & 16.11 & 21.75 & 4.78  & 4.63  &  9.96     & 6.92 & 15.05 & 29.49 & 38.73 & 7.19 & 6.92 &   17.38  \\
	SHS-Net~\cite{li2023shsnet}  	             & 3.95  & 8.55  & 16.13 & 21.53 & 4.91  & 4.67  &  9.96     & 7.41 & 15.34 & 29.33 & 38.56 & 7.74 & 7.28 &   17.61  \\
	\midrule
	\textbf{Unsupervised} &&&&&&&&&&&&&& \\
	Boulch \etal~\cite{boulch2012fast} 	         & 11.80 & 11.68 & 22.42 & 35.15 & 13.71 & 12.38 &  17.86    & 19.00 & 19.60 & 36.71 & 50.41 & 20.20 & 17.84 &   27.29  \\
	PCV~\cite{zhang2018multi} 	                 & 12.50 & 13.99 & 18.90 & 28.51 & 13.08 & 13.59 &  16.76    & 21.82 & 22.20 & 31.61 & 46.13 & 20.49 & 19.88 &   27.02  \\
	Jet~\cite{cazals2005estimating}              & 12.35 & 12.84 & \textbf{18.33} & 27.68 & 13.39 & 13.13 &  16.29    & 20.11 & 20.57 & 31.34 & 45.19 & 18.82 & 18.69 &   25.79  \\
	PCA~\cite{hoppe1992surface} 	             & 12.29 & 12.87 & 18.38 & 27.52 & 13.66 & 12.81 &  16.25    & 19.90 & 20.60 & 31.33 & 45.00 & 19.84 & 18.54 &   25.87  \\
	LRR~\cite{zhang2013point} 	                 & 9.63  & 11.31 & 20.53 & 32.53 & 10.42 & 10.02 &  15.74    & 17.68 & 19.32 & 33.89 & 49.84 & 16.73 & 16.33 &   25.63  \\
	Ours    & \textbf{7.89}  & \textbf{9.85}  &        {18.62} & \textbf{24.89} & \textbf{9.21}  & \textbf{9.29}  &   \textbf{13.29}
			& \textbf{13.74} & \textbf{16.51} & \textbf{31.05} & \textbf{40.68} & \textbf{13.95} & \textbf{13.17} &   \textbf{21.52}    \\
	\bottomrule
\end{tabular} }
\vspace{-0.4cm}
\end{table*}

\section{Experiments}

\noindent\textbf{Architecture}.
To learn the neural gradient function, we use a simple neural network similar to~\cite{atzmon2020sal,ma2020neural,gropp2020implicit}.
It is composed of eight linear layers and a skip connection from the input to the intermediate layer.
Except for the last layer, each linear layer contains $512$ hidden units and is equipped with a ReLU activation function.
The last layer outputs the signed distance for each point.
The parameters of the linear layer are initialized using the geometric initialization~\cite{atzmon2020sal}.
After optimization, we use the network to derive pointwise gradients from the data $P$ (see Fig.~\ref{fig:net}).

\begin{wrapfigure}{l}{0.73\textwidth}
  \vspace{-0.1cm}
	\centering
	\includegraphics[width=\linewidth]{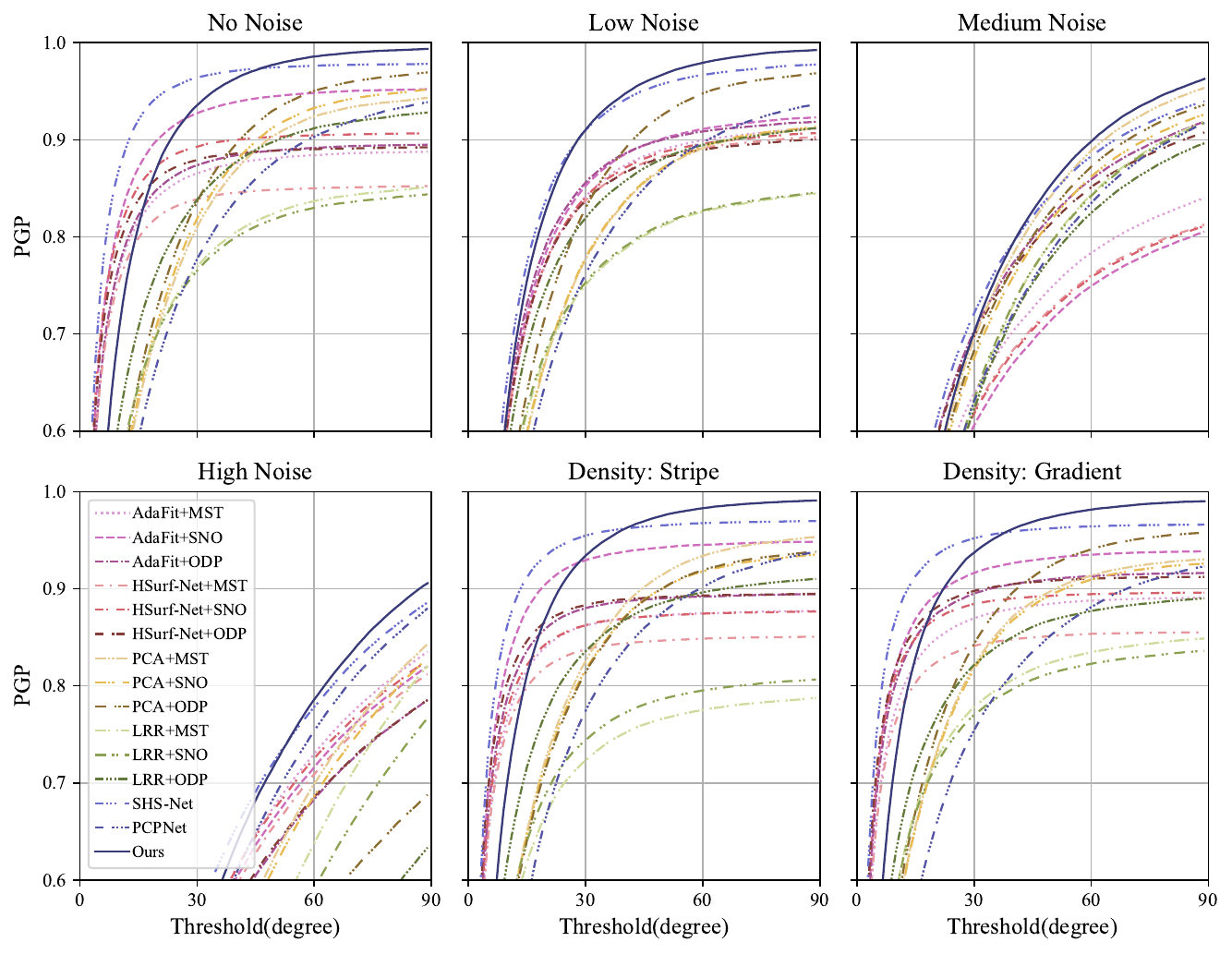}  \vspace{-0.5cm}
  \caption{
    The PGP curves of oriented normal on the FamousShape dataset.
    It is plotted by the percentage of good point normals (PGP) for a given angle threshold.
    Our method has the best value at large thresholds.
  }
  \label{fig:PGP_FamousShape}
  \vspace{-0.2cm}
\end{wrapfigure}

\noindent\textbf{Distribution $\mathcal{D}$}.
We make the distribution $\mathcal{D}$ concentrate in the neighborhood of $P$ in 3D space.
Specifically, it is set by uniform sampling points $p$ from $P$ and placing an isotropic Gaussian $N(p,\sigma^2)$ centered at each $p$, with standard deviation parameter $\sigma$ is adaptively set to the distance from the $l$-th nearest point to $p$~\cite{atzmon2020sal,atzmon2021sald}.
We set $l$ to 25 for the PCPNet dataset and 50 for the FamousShape dataset.

\noindent\textbf{Implementation}.
We choose the neighborhood scale set $\{K_s\}^{N_K}_{s=1}$ as $\{1,K/2,K\}, K \!=\! 8$, and set the parameter $\rho$ of adaptive weight to $60$.
We select $N_Q\!=\!5000$ points from the distribution $\mathcal{D}$ and $N_G\!=\!2500$ points from $P$ to form the input $\{Q, G\}$.
The number of steps is set to $\mathcal{I} \!=\! 2$ for all evaluations.
As for metric, we use normal angle Root Mean Squared Error (RMSE) to evaluate the estimated normals and Percentage of Good Points (PGP) to show the normal error distribution~\cite{li2022hsurf,li2023shsnet}.

\begin{figure}[h]
	\centering
	\includegraphics[width=\linewidth]{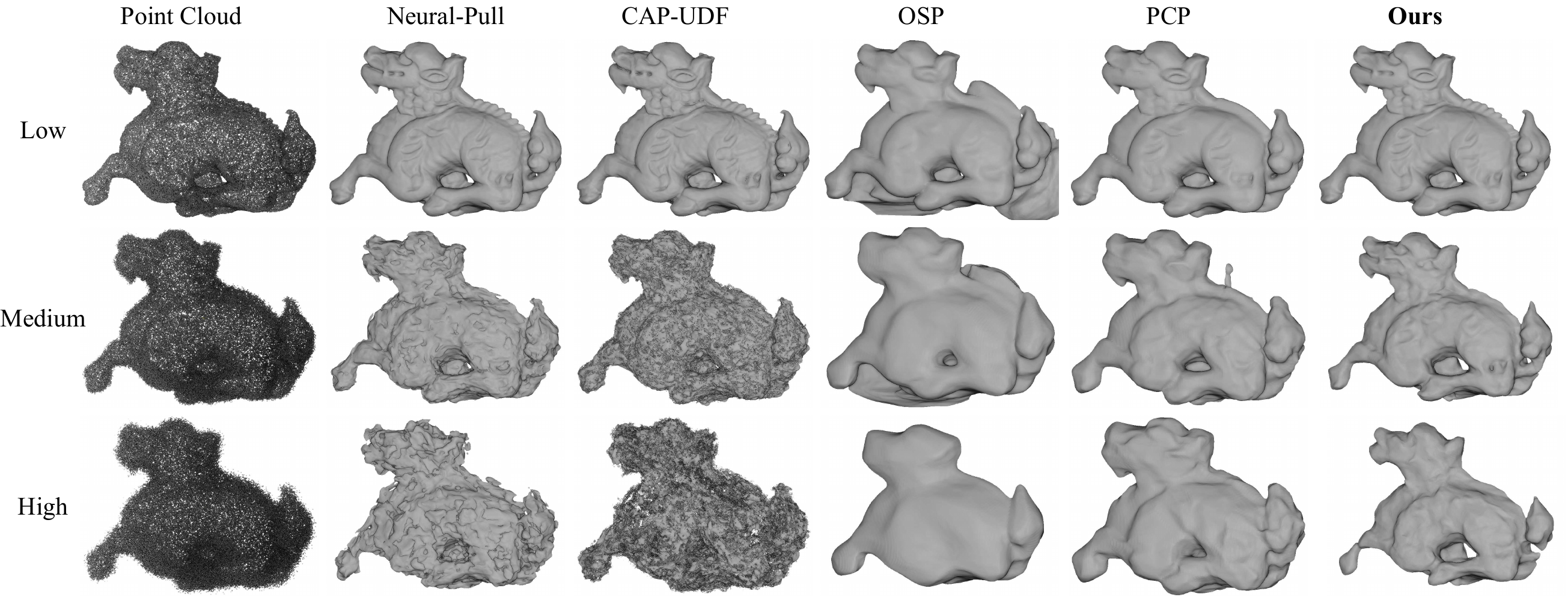}  \vspace{-0.7cm}
  \caption{
    Comparison with surface reconstruction methods using point movement strategy, \eg, Neural-Pull~\cite{ma2020neural}, CAP-UDF~\cite{zhou2022learning}, OSP~\cite{ma2022reconstructing} and PCP~\cite{ma2022surface}.
    The input point clouds have different levels of noise (low, medium and high).
  }
  \label{fig:compRecon_netsuke}
  \vspace{-0.2cm}
\end{figure}

\begin{figure}
	\centering
	\includegraphics[width=\linewidth]{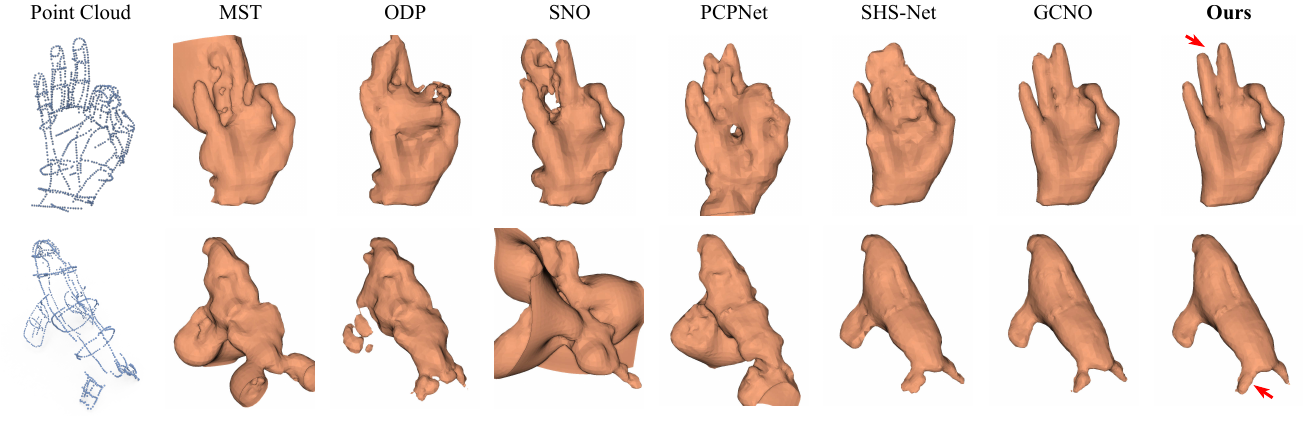}  \vspace{-0.7cm}
  \caption{
    The reconstructed surfaces on wireframe point clouds with sparse and non-uniform sampling.
    There are less than $1000$ points in each input point cloud.
    The initial normals of MST, ODP and SNO are provided by PCA.
    GCNO~\cite{xu2023globally} performs similarly to our method.
  }
  \label{fig:WireframePC}
  \vspace{-0.3cm}
\end{figure}

\subsection{Performance Evaluation}

\noindent\textbf{Evaluation of Oriented Normal}.
In this evaluation, we not only compare with one-stage baseline methods, including PCPNet~\cite{guerrero2018pcpnet}, DPGO~\cite{wang2022deep} and SHS-Net~\cite{li2023shsnet},
but also compare with two-stage baseline methods that combine representative algorithms based on different design philosophies.
Specifically, we use the various combinations of unoriented normal estimation methods (PCA~\cite{hoppe1992surface}, AdaFit~\cite{zhu2021adafit} and HSurf-Net~\cite{li2022hsurf}) and normal orientation methods (MST~\cite{hoppe1992surface}, SNO~\cite{schertler2017towards} and ODP~\cite{metzer2021orienting}).
Note that PCPNet, DPGO, SHS-Net, AdaFit and HSurf-Net rely on supervised training with ground truth normals.
We report quantitative evaluation results on datasets PCPNet and FamousShape~\cite{li2023shsnet} in Table~\ref{table:pcpnet_famousShape_o}.
Our method has better performance under the vast majority of data categories (noise levels and density variations), and achieves the best average results.
It is worth mentioning that our method achieves a large performance improvement on the FamousShape dataset, which contains different shapes with more complex geometries than the PCPNet dataset.
We provide a visual comparison of the estimated normals in Fig.~\ref{fig:error_gradient}, which shows normal vectors of a column section with a gear-like structure.
In Fig.~\ref{fig:error_map}, we show the normal error map on point clouds with uneven sampling and multi-branch structure.
The normal error distribution is shown in Fig.~\ref{fig:PGP_FamousShape}.
It is clear that our unsupervised method has advantages over baselines in all categories when the angle threshold is larger than $50^{\circ}$.
The evaluation results demonstrate that our method can consistently provide good oriented normals in the presence of various complex structures and density variations.

\noindent\textbf{Evaluation of Unoriented Normal}.
In general, estimating unoriented normals is easier than estimating oriented normals as it does not need to explore more information to determine the orientation, but only focuses on finding the perpendicular of a local plane or surface according to the input patch.
Moreover, existing deep learning-based methods for unoriented normal estimation all rely on supervised training with ground truth normals.
In contrast, our method is designed for estimating oriented normal in an unsupervised manner.
To evaluate the unoriented normals, we compare with the baselines using our oriented normals and ignore the orientation of normals.
In Table~\ref{table:pcpnet_famousShape}, we report the quantitative evaluation results of supervised and unsupervised methods on datasets PCPNet and FamousShape.
We can see that, compared to unsupervised methods, our method achieves significant performance gains on almost all data categories of these two datasets and has the best average results.
The supervised methods learn from ground truth normals, and their supervision provides perfect surface perpendiculars, leading to superior results.
Our better results than PCPNet highlight the strong learning ability of our method without labels.


\begin{wraptable}{r}{0.43\textwidth}
    \footnotesize
    \centering
    \setlength{\tabcolsep}{1.2mm}
    \vspace{-0.65cm}
    \caption{
        Comparison of oriented normal estimation. We use implicit representation methods to estimate oriented normals.
    }
    \vspace{0.1cm}
    \resizebox{\linewidth}{!}{
    \begin{tabular}{lccc}
    \toprule
        & Noise    & Density   & Average  \\
    \midrule
        IGR~\cite{gropp2020implicit}       & 54.77           & 75.90           & 65.33            \\
        SAP~\cite{peng2021shape}           & 57.56           & 41.32           & 49.44            \\
        SAL~\cite{atzmon2020sal}           & 46.69           & 43.78           & 45.24            \\
        Neural-Pull~\cite{ma2020neural}    & 48.48           & 26.22           & 37.35            \\
        Ours                               & \textbf{36.92}  & \textbf{26.08}  & \textbf{31.50}   \\
    \bottomrule
    \end{tabular}}
    \label{tab:sdf_methods}
    \vspace{-0.1cm}
\end{wraptable}

\begin{wrapfigure}{r}{0.53\textwidth}
  \vspace{-.7cm}
	\centering
	\includegraphics[width=\linewidth]{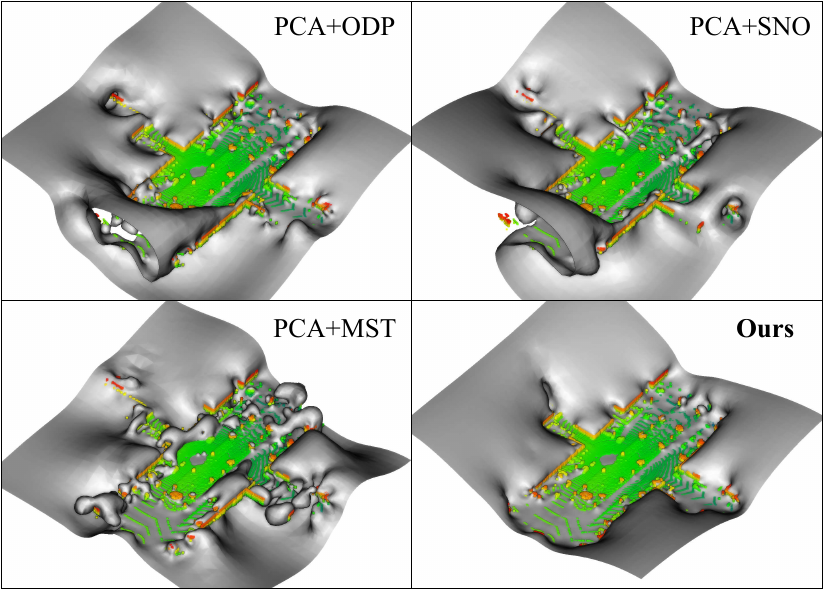}  \vspace{-0.6cm}
  \caption{
    The reconstructed surfaces on the KITTI dataset using estimated normals of unsupervised methods.
    Points are colored with their heights.
  }
  \label{fig:compRecon_KITTI}
  \vspace{-0.4cm}
\end{wrapfigure}

\noindent\textbf{Surface Reconstruction}.
For surface reconstruction from point clouds, we compare our method with the latest methods in two different ways.
(1) We determine the zero level set of the learned implicit function $f$ via signed distances, and use the marching cubes algorithm~\cite{lorensen1987marching} to extract the global surface representation of the point cloud.
A visual comparison of the extracted surfaces on point clouds with different noise levels is shown in Fig.~\ref{fig:compRecon_netsuke}, where our surfaces are much cleaner and more complete under the interference of noise.
(2) Based on the estimated oriented normals, we employ the Poisson reconstruction algorithm~\cite{kazhdan2013screened} to generate surfaces from wireframe point clouds, and the results in Fig.~\ref{fig:WireframePC} show that our method can handle extremely sparse and uneven data.
The reconstructed surfaces of unsupervised methods on LiDAR point clouds of the KITTI dataset~\cite{geiger2012we} are shown in Fig.~\ref{fig:compRecon_KITTI}, where our undistorted surface exhibits a more realistic road scene of real-world data.

Moreover, we repurpose some implicit representation methods, which are designed for surface reconstruction, to estimate oriented normals from point clouds.
The comparison of normal RMSE on some selected point clouds in the datasets PCPNet and FamousShape is reported in Table~\ref{tab:sdf_methods}.
A visual comparison of the reconstructed surfaces on noisy point clouds is shown in the supplementary material.
We can see that our method has clear advantages, especially on noisy point clouds.
We provide more evaluation results on different data in the supplementary material.

\begin{table*}[t]
\centering
\footnotesize
\setlength{\tabcolsep}{1.0mm}
\caption{
    Ablation studies for unoriented and oriented normal estimation on the FamousShape dataset.
}
\label{tab:ablation}
\resizebox{\linewidth}{!}{
\begin{tabular}{ll|cccc|cc| >{\columncolor{mygray}} c||  cccc|cc| >{\columncolor{mygray}} c}
    \toprule
    & \multirow{3}{*}{Category} & \multicolumn{7}{c||}{\textbf{Unoriented Normal}} & \multicolumn{7}{c}{\textbf{Oriented Normal}} \\
    \cmidrule(r){3-16}
    &
    & \multicolumn{4}{c|}{Noise} & \multicolumn{2}{c|}{Density} &    & \multicolumn{4}{c|}{Noise} & \multicolumn{2}{c|}{Density} &  \\
    &
    & None & 0.12\% & 0.6\% & 1.2\% & Stripe & Gradient & \multirow{-2}{*}{{Average}}
    & None & 0.12\% & 0.6\% & 1.2\% & Stripe & Gradient & \multirow{-2}{*}{{Average}} \\
    \midrule
    \multirow{6}{*}{\textbf{(a)}}
    & w/o $\mathcal{L}_{reg}$
    & 13.76 & 16.45 & 31.32 & 42.56 & 13.72 & 12.87 &   21.78      & 16.63 & 19.15 & 36.77 & 55.22 & 16.89 & 16.35 &   26.84  \\
    & w/o $\mathcal{L}_d$
    & 13.98 & 16.66 & 30.88 & 40.70 & 14.01 & 13.30 &   21.59      & 16.74 & 19.30 & 35.72 & 49.98 & 17.21 & 17.39 &   26.06  \\
    & w/o $\mathcal{L}_{con}$
    & 13.93 & 17.13 & 33.81 & 45.95 & 13.95 & 13.02 &   22.97      & 23.97 & 26.70 & 40.49 & 69.06 & 23.26 & 17.87 &   33.56  \\
    & w/o $\mathcal{L}_v$
    & 28.41 & 31.34 & 49.04 & 49.65 & 30.60 & 26.47 &   35.92      & 46.97 & 43.00 & 73.64 & 75.61 & 43.10 & 35.33 &   52.94  \\
    & w/o $w$
    & 13.76 & 16.56 & 31.45 & 41.61 & 14.00 & 12.83 &   21.70      & 16.81 & 19.40 & 37.06 & 52.97 & 18.12 & 16.11 &   26.74  \\
    & w/ $\mathcal{L}_v \cdot w$
    & 13.44 & 16.14 & 31.35 & 41.80 & 13.71 & 12.90 &   21.56      & 16.20 & 18.99 & 36.66 & 51.47 & 17.61 & 18.60 &   26.59  \\
    \hline
    \multirow{1}{*}{\textbf{(b)}}
    & w/o $G$
    & 13.86 & 16.73 & 31.10 & 40.73 & 14.02 & 13.30 &   21.62      & 16.99 & 21.66 & 36.70 & 50.30 & 17.37 & 20.35 &   27.23  \\
    \hline
    \multirow{2}{*}{\textbf{(c)}}
    & $\mathcal{I}=1$
    & 14.03 & 17.05 & 33.42 & 45.97 & 13.94 & 13.36 &   22.96      & 23.39 & 26.34 & 40.10 & 63.89 & 19.47 & 27.89 &   33.51  \\
    & $\mathcal{I}=3$
    & 13.94 & 16.61 & 30.63 & 39.82 & 13.88 & 13.06 &   21.32      & 16.91 & 19.69 & 35.65 & 50.08 & 17.64 & 16.46 &   26.07  \\
    \hline
    \multirow{6}{*}{\textbf{(d)}}
    & $\{1\}$
    & 14.11 & 17.25 & 31.62 & 41.36 & 13.72 & 12.85 &   21.82      & 23.87 & 23.14 & 36.47 & 53.32 & 17.82 & 16.35 &   28.50  \\
    & $\{K\}$
    & 14.31 & 16.67 & 30.43 & 40.12 & 14.45 & 13.37 &   21.56      & 17.26 & 19.36 & 35.34 & 49.13 & 18.33 & 16.88 &   26.05  \\
    & $\{1,K\}$
    & 13.72 & 16.62 & 30.77 & 40.28 & 13.92 & 13.06 &   21.39      & 16.54 & 19.40 & 35.65 & 51.98 & 17.02 & 16.55 &   26.19  \\
    & $K=4$
    & 13.60 & 17.34 & 30.90 & 40.43 & 13.70 & 12.56 &   21.42      & 16.66 & 23.60 & 36.17 & 49.44 & 16.86 & 14.72 &   26.24  \\
    & $K=16$
    & 14.32 & 16.69 & 30.86 & 41.25 & 14.40 & 13.40 &   21.82      & 17.26 & 19.42 & 35.85 & 50.98 & 17.43 & 17.17 &   26.35  \\
    & $K=32$
    & 15.02 & 17.18 & 30.61 & 41.15 & 15.44 & 14.26 &   22.28      & 17.82 & 19.79 & 35.70 & 50.64 & 19.28 & 18.22 &   26.91  \\
    \hline
    & \multirow{1}{*}{\textbf{Final}}
    & {13.74} & {16.51} & {31.05} & {40.68} & {13.95} & {13.17} &   {21.52}
    & {16.57} & {19.28} & {36.22} & {50.27} & {17.23} & {17.38} &   {26.16}  \\
    \bottomrule
\end{tabular} }
\vspace{-0.4cm}
\end{table*}

\subsection{Ablation Studies}  \label{sec:ablation}

We aim to achieve the best performance on average in both oriented and unoriented normal estimation.
We provide the ablation results on the FamousShape dataset in Table~\ref{tab:ablation} (a)-(d), which are discussed as follows.

\noindent\textbf{(a) Loss}.
We do not use one of the four loss terms in Eq.~\eqref{eq:loss} and the weight $w$ in Eq.~\eqref{eq:loss_con}, respectively.
We also try incorporating the weight $w$ into $\mathcal{L}_{v}(\bm{\theta})$ of Eq.~\eqref{eq:loss_grad}.
These ablations do not provide better performance in both oriented and unoriented normal estimation.
We observe that the performance drops a lot without using $\mathcal{L}_{v}(\bm{\theta})$, and that $\mathcal{L}_{con}(\bm{\theta})$ has a larger impact on performance.
Note that $\mathcal{L}_{d}(\bm{\theta})$ plays a more important role on the PCPNet dataset than the FamousShape dataset in our experiments.

\noindent\textbf{(b) Input}.
We do not sample a subset $G$ from $P$, and take $Q$ as the input instead of $\{Q, G\}$.
The results verify the positive effect of $G$ on explicitly indicating the surface and improving performance.

\noindent\textbf{(c) Iteration}.
We set the number of steps $\mathcal{I}$ to $1$ and $3$, respectively.
More iterations bring better performance, but we still choose $\mathcal{I} \!=\! 2$ after weighing the running time and memory usage.

\noindent\textbf{(d) Scale Set}.
The scale set $\{K_s\}^{N_K}_{s=1}$ in Eq.~\eqref{eq:scale} is chosen to be $\{1,K/2,K\}, K \!=\! 8$ in our implementation.
Here we change the scale set to different sizes, \eg, $\{1\}$, $\{K\}$ and $\{1,K\}, K \!=\! 8$, or different values of $K$, \eg, 4, 16 and 32.
Some of these settings have advantages in a single evaluation, but do not give better results for both oriented and unoriented normal estimation, such as $\{K\}$ and $\{1,K\}$.

\section{Conclusion}
\vspace{-0.2cm}

In this work, we propose to learn neural gradient functions from point clouds to estimate oriented normals without requiring ground truth normals.
Specifically, we introduce loss functions to facilitate query points to iteratively reach the moving targets and aggregate onto the approximated surface, thereby learning a global surface representation of the data.
Meanwhile, we incorporate gradients into the surface approximation to measure the minimum signed deviation of queries, resulting in a consistent gradient field associated with the surface.
Extensive evaluation and ablation experiments are provided, and our excellent results in both unoriented and oriented normal estimation demonstrate the effectiveness of innovations in our design.
Future work includes adding more local constraints to further improve the accuracy of normals.

\section{Acknowledgement}
This work was supported by National Key R\&D Program of China (2022YFC3800600), the National Natural Science Foundation of China (62272263, 62072268), and in part by Tsinghua-Kuaishou Institute of Future Media Data.


{\small
\bibliographystyle{ieee_fullname}
\bibliography{egbib}
}



\section*{Supplementary material}

# NeuralGF: Unsupervised Point Normal Estimation by Learning Neural Gradient Function — Supplementary Material —

Qing Li[1]  Huifang Feng[2]  Kanle Shi[3]  Yue Gao[1]  Yi Fang[4]
Yu-Shen Liu[1]*  Zhizhong Han[5]
[1]School of Software, Tsinghua University, Beijing, China
[2]School of Informatics, Xiamen University, Xiamen, China
[3]Kuaishou Technology, Beijing, China
[4]Center for Artificial Intelligence and Robotics, New York University Abu Dhabi, Abu Dhabi, UAE
[5]Department of Computer Science, Wayne State University, Detroit, USA
{leoqli, gaoyue, liuyushen}@tsinghua.edu.cn  fenghuifang@stu.xmu.edu.cn
yfang@nyu.edu  h312h@wayne.edu

## 1 Evaluation Metrics

As in [7, 13, 12], we use the Root Mean Squared Error (RMSE) to evaluate the unoriented and oriented normal results of different methods. Specifically, the normal RMSE is calculated as the vector angles between the ground-truth normals $\hat{\mathbf{n}}_i$ and the estimated normals $\mathbf{n}_i$, that is,

$$\text{RMSE}_U = \sqrt{\frac{1}{I}\sum_{i=1}^{I}\big(\arccos(|\hat{\mathbf{n}}_i \odot \mathbf{n}_i|)\big)^2}\ , \tag{1}$$

$$\text{RMSE}_O = \sqrt{\frac{1}{I}\sum_{i=1}^{I}\big(\arccos(\hat{\mathbf{n}}_i \odot \mathbf{n}_i)\big)^2}\ , \tag{2}$$

where $\text{RMSE}_U$ and $\text{RMSE}_O$ are metrics for evaluating unoriented and oriented normal results, respectively. The angular error range in unoriented normal evaluation is between $0°$ and $90°$, while the angular error range in oriented normal evaluation is between $0°$ and $180°$. $I$ is the number of point normals to be evaluated. $\odot$ indicates the inner product of two normal vectors and $|\cdot|$ denotes the absolute value.

In addition, to analyze the error distribution of the normal results, we also use the Percentage of Good Points (PGP) [27, 13, 12], which describes the percentage of good points whose normal angle errors are smaller than several specific angle thresholds. Specifically, the PGP with respect to a threshold $\tau$ is computed by

$$\text{PGP}_U(\tau) = \frac{1}{I}\sum_{i=1}^{I}\mathcal{H}\big(\arccos(|\hat{\mathbf{n}}_i \odot \mathbf{n}_i|) < \tau\big), \tag{3}$$

$$\text{PGP}_O(\tau) = \frac{1}{I}\sum_{i=1}^{I}\mathcal{H}\big(\arccos(\hat{\mathbf{n}}_i \odot \mathbf{n}_i) < \tau\big), \tag{4}$$

where $\text{PGP}_U(\tau)$ and $\text{PGP}_O(\tau)$ are metrics for evaluating unoriented and oriented normal results, respectively. $\mathcal{H}$ represents an indicator function that measures whether the normal angle error is smaller than the given threshold $\tau$.

---

*Corresponding author

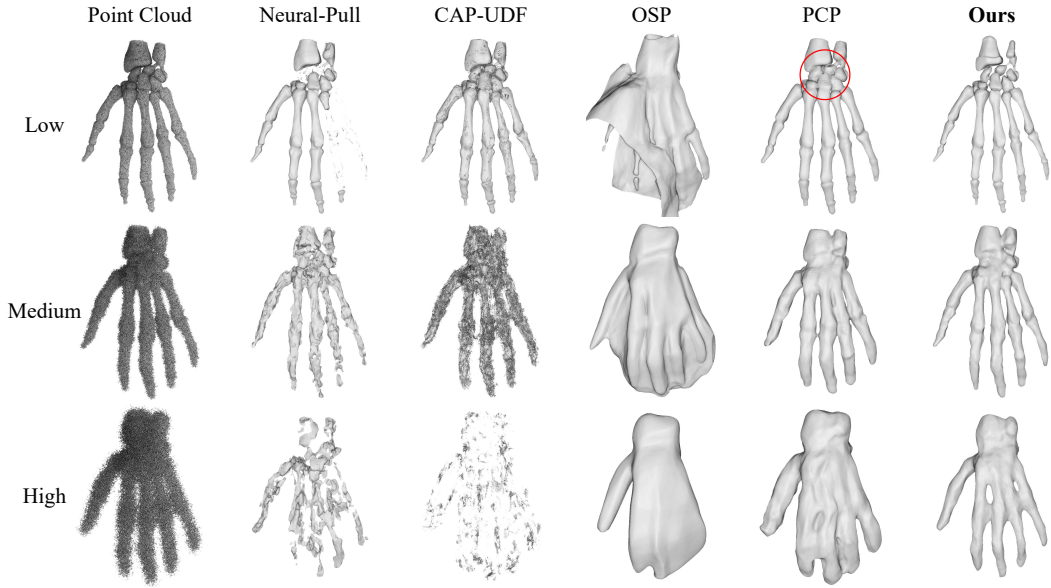

Figure 1: Comparison with surface reconstruction methods using point movement strategy, including Neural-Pull [17], CAP-UDF [26] OSP [18] and PCP [19]. The input point clouds have different levels of noise (low, medium and high).

Table 1: The average normal angle RMSE, number of learnable network parameters, and running time of learning-based oriented normal estimation methods on the PCPNet dataset. We provide optimization time (*i.e.*, training time in the bracket) and inference time of our method. We only provide inference times for other learning-based methods since we directly use their pre-trained network models.

|  | PCPNet [7] | HSurf-Net [13] +ODP [20] | AdaFit [27] +ODP [20] | SHS-Net [12] | Ours |
| --- | --- | --- | --- | --- | --- |
| RMSE | 37.66 | 31.07 | 30.93 | 19.79 | **18.70** |
| Time (seconds per 100k points) | 63.02 | 72.47+236.35 | 56.23+248.54 | 65.89 | (1257.37)+**0.10** |
| Parameters (million) | 22.36 | 2.16+0.43 | 4.87+0.43 | 3.27 | **0.46** |
| Relative parameter | **49×** | **6×** | **12×** | **7×** | **1×** |

## 2 Baseline Methods

In the evaluation of unoriented normal estimation, the baseline methods include three types: (1) The traditional normal estimation methods, such as PCA [8], Jet [4] and LRR [25]; (2) The learning-based surface fitting methods, such as DeepFit [2], AdaFit [27] and GraphFit [11]; (3) The learning-based normal regression methods, such as PCPNet [7], Nesti-Net [3], NeAF [14], HSurf-Net [13] and SHS-Net [12]. In the evaluation of oriented normal estimation, the baseline methods include two types: (1) The two-stage based methods that are built by combining three unoriented normal estimation methods (PCA [8], AdaFit [27] and HSurf-Net [13]) and three normal orientation methods (MST [8], ODP [20] and SNO [22]), such as PCA+MST, AdaFit+SNO and HSurf-Net+ODP. (2) The end-to-end based oriented normal estimation methods, such as PCPNet [7] and SHS-Net [12].

## 3 Complexity and Efficiency

In oriented normal evaluation, SHS-Net, PCPNet, HSurf-Net, AdaFit and ODP are learning-based methods, and the others are traditional methods. In this evaluation experiment, we compare our method with the learning-based methods on a machine equipped with NVIDIA 2080 Ti GPUs. As shown in Table 1, we report the average normal angle RMSE, the number of the learnable network parameters, and the running time of each method for oriented normal estimation on the PCPNet dataset [7]. Our method improves the state-of-the-art results while using much fewer parameters. We

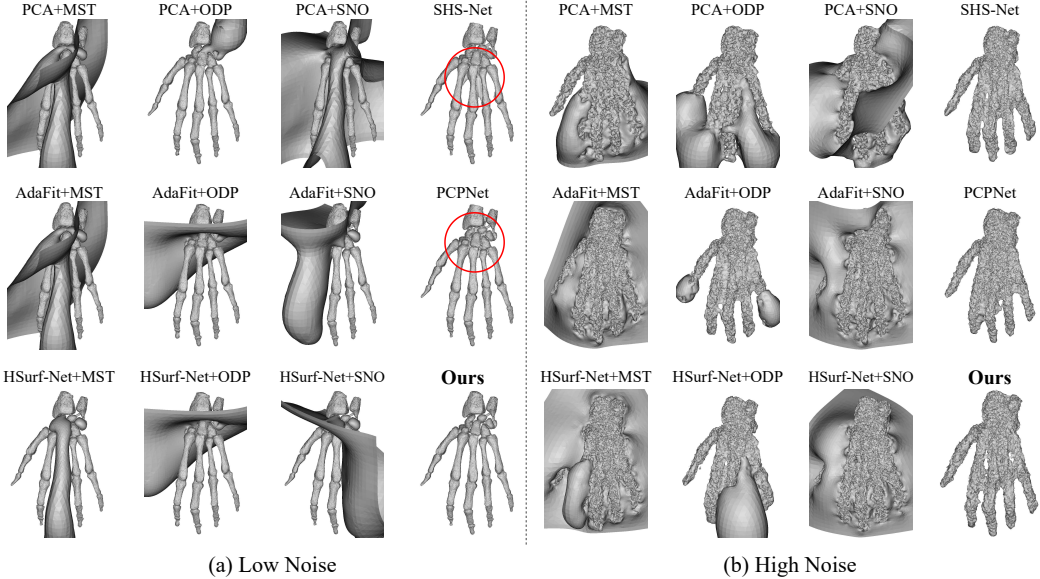

(a) Low Noise  (b) High Noise

Figure 2: Comparison of surface reconstruction results using oriented normals estimated by different methods. The surfaces are reconstructed from point clouds with low noise (a) and high noise (b).

provide the optimization time and inference time of our method, where the optimization time is long and the inference time is very short.

Note that we only report the running time of various methods for predicting normal from point cloud (*i.e.*, testing), and Table 1 does not include the training time of other methods. The other supervised methods can be trained in advance using ground truth, and then their trained models are used for testing. In contrast, our unsupervised method does not require training data and ground truth to train a model, but requires optimization for each shape in the test data to obtain its learned function, so we provide our optimization (*i.e.*, training) time in the table. As for training time, our method has a similar time cost (about 40 hours) to the SOTA method SHS-Net on the entire PCPNet training dataset using an NVIDIA 2080 Ti GPU.

## 4  Surface Reconstruction

We demonstrate the global surface learning and normal estimation capabilities of our method, and verify its reliability by applying the estimated normals to downstream applications, such as surface reconstruction from point clouds. We compare our method with the state-of-the-art methods in two different ways as follows.

**Comparison with Surface Reconstruction Methods**. We determine the zero level set of the learned implicit function $f$ via signed distances, and use the marching cubes algorithm [16] to extract the global surface representation of the point cloud. To evaluate the reconstructed surfaces, we compare our method with baseline methods that are designed for surface reconstruction using point movement strategy, including Neural-Pull [17], CAP-UDF [26], OSP [18] and PCP [19]. A visual comparison of the extracted surfaces on point clouds with different noise levels is shown in Fig. 1. We can see that our reconstructed surfaces have more accurate structures compared to the baseline methods.

**Comparison with Oriented Normal Estimation Methods**. Based on the estimated oriented normals, we employ the Poisson reconstruction algorithm [10] to generate surfaces from point clouds. In Fig 2, we show the reconstructed surfaces on point clouds with different noise levels. In Fig 3, we demonstrate the reconstructed surfaces on point clouds with complex geometries. Compared to baseline methods, the Poisson algorithm can reconstruct more complete geometries with details from various point cloud data based on the accurate normals estimated by our method. In Fig 4, we show the reconstructed surfaces of unsupervised methods on LiDAR point clouds of the KITTI dataset [5], which is captured in outdoor real-world road scenes. The results show that our undistorted surface

3

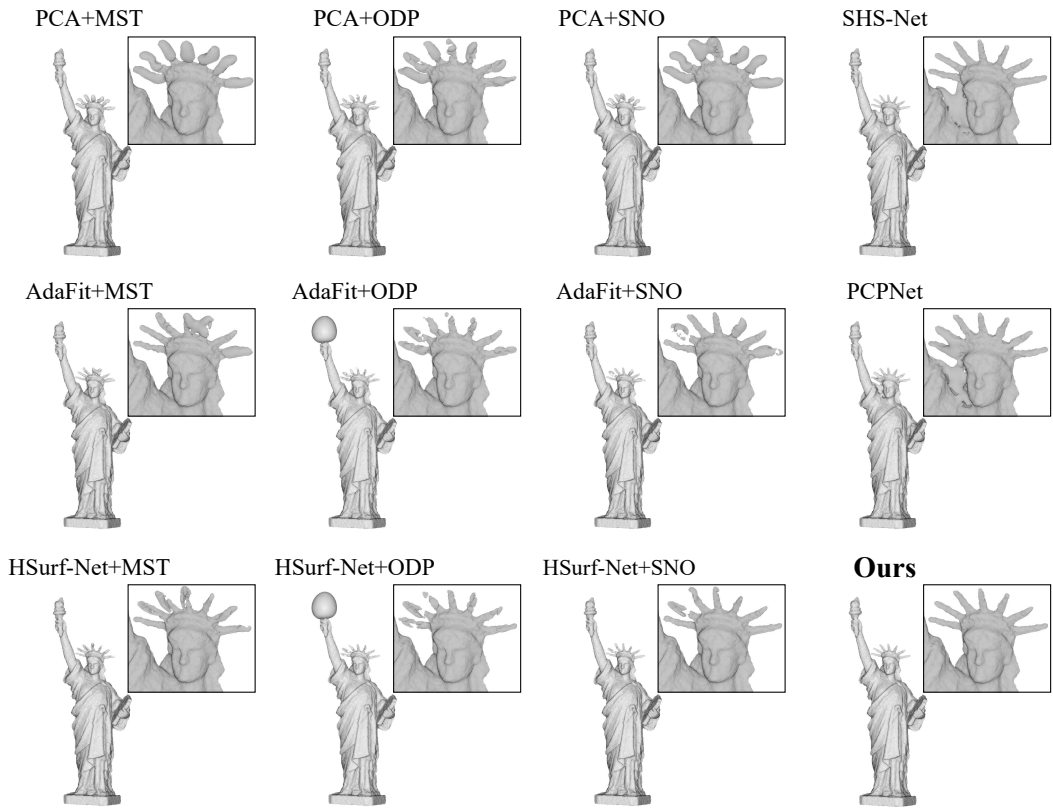

Figure 3: Comparison of surface reconstruction results using oriented normals from different methods. A partially enlarged view is provided for each shape.

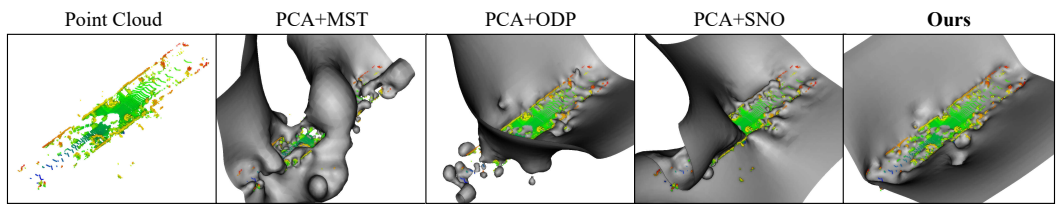

Figure 4: The reconstructed road surfaces on the KITTI dataset using estimated normals of unsupervised methods. Points are colored with their heights.

exhibits a more realistic road scene of real-world data. The comparative results in our experiments demonstrate that our method can accurately estimate oriented normals on point clouds with noise, density variations, and complex geometries, thereby facilitating downstream tasks.

## 5 Limitation, Failure Cases and Broader Impact

Generally, estimating oriented normals with consistent orientations is more challenging than estimating unoriented normals, *i.e.*, finding the perpendiculars of local planes, which has been well-studied over the past decades. The most recent approaches, such as AdaFit and HSurf-Net, can learn to estimate accurate unoriented normal results from point clouds in a supervised manner, but their global orientation consistency can only be achieved by employing a post-processing step, such as MST-based orientation propagation [8]. As we observed in the evaluation experiments, the bottleneck of estimating high-precision oriented normals is correctly determining the normal orientation. Even if the corresponding unoriented normal is very accurate, it is not guaranteed to get a good oriented normal.

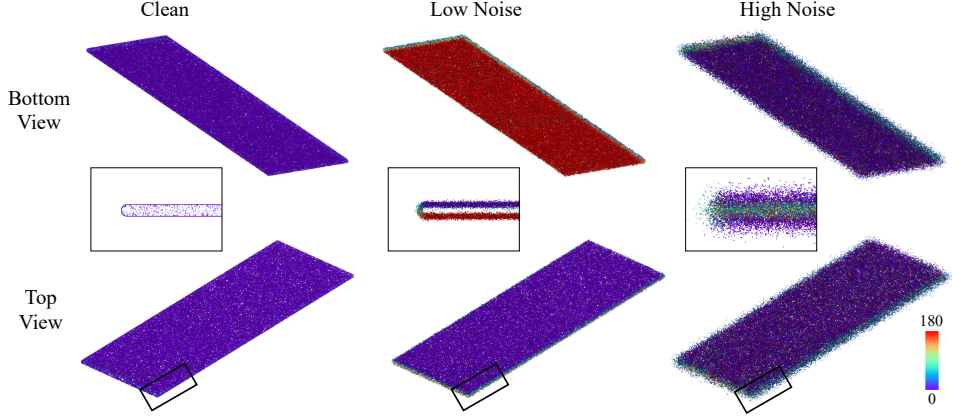

Figure 5: Visualization of oriented normal errors on point clouds with sheet structures, which are taken from the PCPNet dataset [7]. Our method can accurately estimate oriented normals on the clean point cloud, but fail on the point clouds with noise. The unoriented normals on the low noise point cloud are still accurate (shown in red and purple), while the high noise point cloud has many normals with wrong directions (shown in green and yellow). We map the normal angle error to RGB colors. We provide the bottom view, top view, and locally enlarged view of the shape.

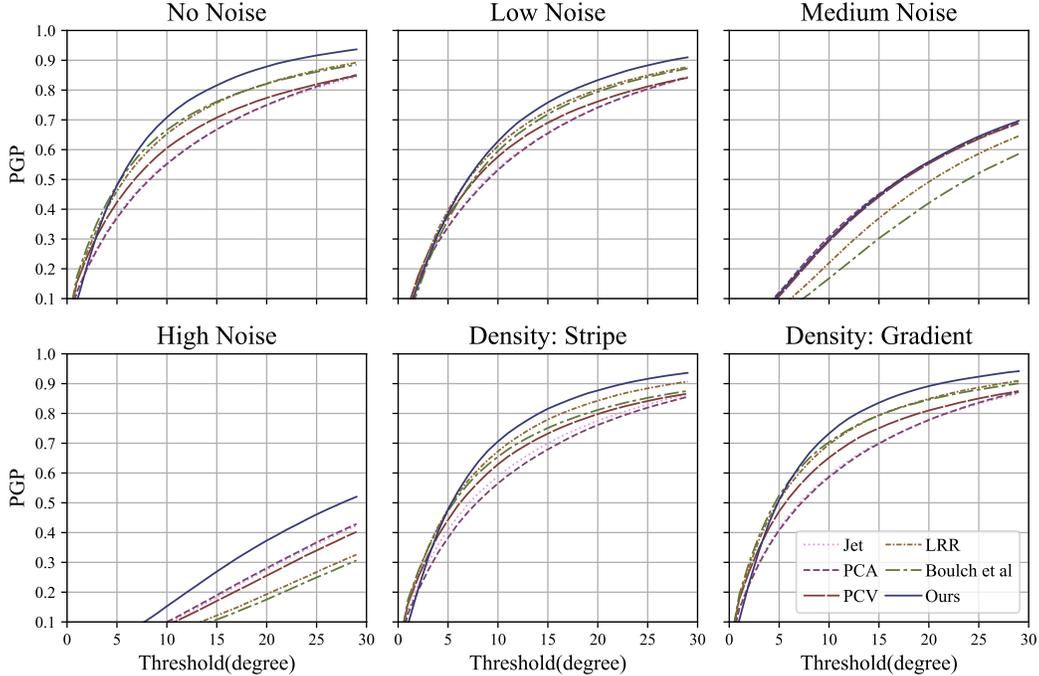

Figure 6: Unoriented normal PGP of unsupervised methods on the FamousShape dataset [12]. The graph is plotted by the percentage of good point normals (PGP) for given angle thresholds. Our method has the best value at large thresholds.

In this work, we propose a novel method to estimate oriented normals directly from 3D point clouds with noise, outliers, and density variations in an unsupervised manner. As we introduced in the paper, our method determines the normal orientation based on learning the neural gradient function, which encourages the neural network to fit global surfaces from the input point clouds and yield consistent unit-norm gradients at each point. In contrast to existing normal estimation methods that use supervised learning of normals to fit surfaces from local patches, the newly proposed approach learns directly from raw data. Our method achieves the state-of-the-art performance while using much fewer network parameters. A limitation of our method is that the neural network needs to fit the model from each single point cloud. The resulting network model cannot be used for shapes not seen during

5

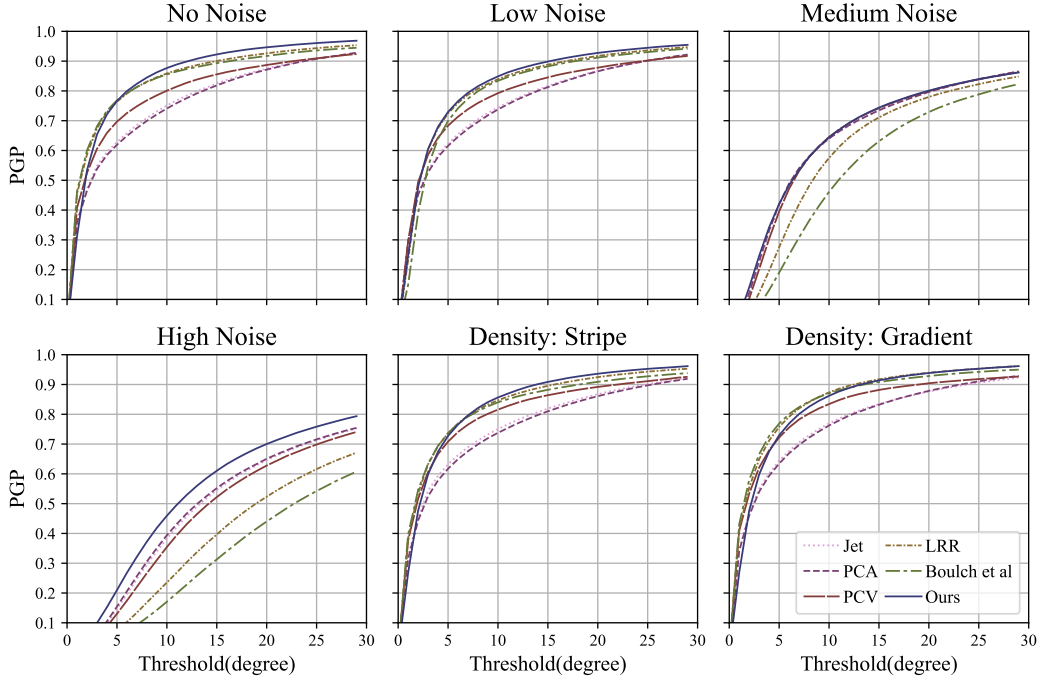

Figure 7: Unoriented normal PGP of unsupervised methods on the PCPNet dataset [7]. The graph is plotted by the percentage of good point normals (PGP) for given angle thresholds. Our method has the best value at large thresholds.

Table 2: Comparison of oriented normal estimation on sparse point clouds.

|  | HSurf-Net [13] +ODP [20] | PCPNet [7] | PCA+ MST [8] | GCNO [24] | SHS-Net [12] | Ours |
|---|---|---|---|---|---|---|
| RMSE | 62.51 | 48.48 | 45.40 | 41.24 | 32.64 | **24.35** |

optimization, and this limitation also exists in some methods of learning implicit functions [17, 26, 19]. Another limitation comes from the construction of query point set $Q$, which is sampled from raw data $P$ according to a distribution $\mathcal{D}$. The default distribution $\mathcal{D}$ in our experiments is suitable for most shapes with various structures. However, we observed that some special structures, such as multi-layer sheets, lead to the inability to estimate oriented normals with consistent orientations. As shown in Fig. 5, we provide some typical cases of these structures. The results shown in the figure are obtained on a clean point cloud and two noisy point clouds. Our method can accurately estimate unoriented normals from these point clouds due to the simple geometry of these shapes. Furthermore, we can see that our method estimates accurate oriented normals on the clean point cloud, but fails on the two noisy point clouds. As shown in the partially enlarged image in the figure, in the clean point cloud, there are clear gaps between the two sheets. In the two noisy point clouds, the points of the two sheets are very close, and even enter into each other's point sets, so that the gap between the two sheets is filled and disappears. This fusion of points will be exacerbated when sampling $Q$ from $P$ using the distribution $\mathcal{D}$, making the model unable to distinguish the inside and the outside of the shape. Eventually, the oriented normals of the two sheets may face the same direction (point cloud with low noise), or even randomly towards the sides of the sheet (point cloud with high noise).

As for the broader impact, various downstream tasks in 3D computer vision can benefit from our method, such as point cloud denoising, surface reconstruction, rendering, segmentation, and so on. The comparison results in Fig. 2-4 demonstrate that our more accurate normals can enhance the performance of the traditional surface reconstruction algorithm. In addition, normal estimation methods are also widely used in the field of robot navigation and grasping, where more reliable normal results can help robots recognize objects and scenes. A potential negative impact is that if the estimated normals are used in some autonomous platforms, it may prevent the controller from making correct decisions when the normal estimator does not provide reliable normals or even fails in some scenarios, resulting in damage or injury.

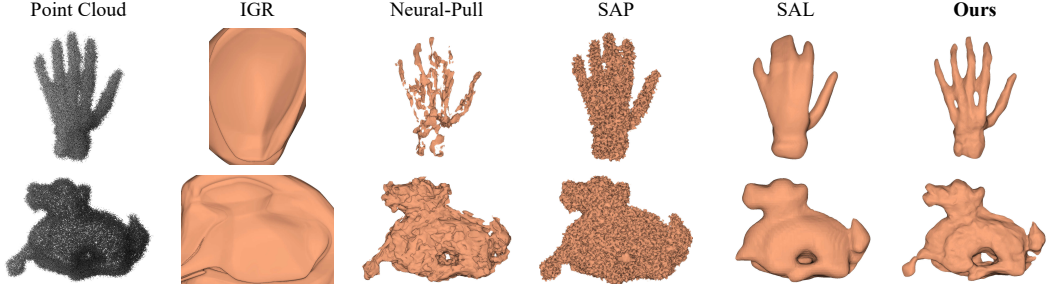

Figure 8: Comparison with implicit representation methods for surface reconstruction, including IGR [6], Neural-Pull [17], SAP [21] and SAL [1]. The input point clouds are noisy.

## 6 More Results

In order to verify the performance on sparse point cloud data, we further conduct an evaluation experiment on a point cloud set that has the same shapes as the FamousShape dataset [12] but each point cloud in this data set has only 5000 points. As shown in Table 2, we provide quantitative comparison results of oriented normal estimation on this sparse point cloud set. It can be observed that our method has the best RMSE result. This evaluation and the comparisons of the challenging cases in the main paper demonstrate the good performance of our method on sparse point clouds.

A visual comparison of the reconstructed surfaces with some implicit representation methods is shown in Fig. 8. In Fig. 6 and Fig. 7, we show the unoriented normal PGP of unsupervised methods on the datasets FamousShape and PCPNet, respectively. The results show that our method has a better performance at the vast majority of thresholds. In Fig. 9, we visualize the position changes of the point cloud after moving through multiple steps in the optimization pipeline. In Fig. 10, we visualize the angle RMSE of the oriented normals estimated by our method on some complicated shapes of the PCPNet dataset [7]. We map the errors to RGB colors to render the point cloud. In Fig. 11- 12, we visualize the angle RMSE of the oriented normals estimated by our method on all shapes of the FamousShape dataset [12].

## 7 More Discussions

### 7.1 Hyperparameter tuning

In all experiments, we use the same network structure and loss weight factors. We did not perform hyperparameter tuning for each shape. We use the same parameters in all categories (clean, noise, and density variation) of a dataset for a fair comparison with other methods. We use the same hyperparameters for all shapes in a benchmark dataset, and only use different parameters to adapt different shapes in different datasets. So, a different input point set does not require different hyperparameters in a dataset and each hyperparameter can be used for a wide range of shapes. Specifically, our method has a hyperparameter needed to be tuned, *i.e.*, determining the standard deviation of distribution $\mathcal{D}$ for different datasets. This hyperparameter is predefined in the code we provide, and we tend to choose a larger value for the dataset with sparse sampling or high curvature.

The hyper-parameter is first set empirically and then tuned according to the experimental results over the validation dataset. The metric we use to tune the hyper-parameter is the RMSE of the oriented normal. Same as existing methods, this metric is also used in evaluation experiments. Specifically, both the PCPNet dataset and the FamousShape dataset contain six categories, we simply choose the average RMSE over the validation dataset as the main indicator when tuning the hyper-parameter for each dataset. As with existing methods, we use standard data splits (training/validation/testing sets) for the datasets used, and the KITTI dataset is only used as the testing set.

### 7.2 Explanation of density-based experiments

The 'gradient' simulates data collected using a 3D scanner, where nearby points are dense while far points are sparse. To achieve this, we give higher weight to points that are closer to the simulated scanner in the probability-based sampling process. The 'stripe' simulates the occlusion situation

7

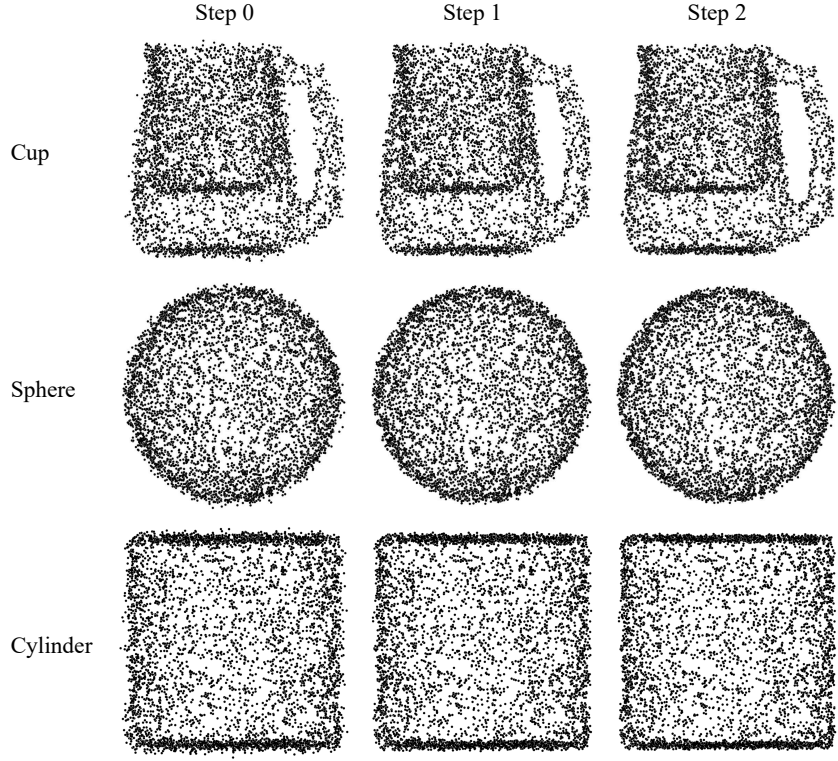

Figure 9: Visualization of the point cloud during the optimization. Our optimization pipeline is formulated as an iterative moving operation of the query point positions. The figure shows the point clouds after moving through multiple steps. During the movement of the second step, the position of the point changes very little. Note that the original input point clouds contain noise and they come from the PCPNet dataset.

during the data collection, making the points in the occluded area sparse or disappear. To achieve that, we divide the shape into multiple areas and sample the points in specific areas with extremely low weights.

### 7.3 Effect of density variation

The optimization of our method is formulated as an iterative moving operation of the input point position. This strategy and the constraints for gradient uniformity of multi-scale neighborhood size make the model have the ability to handle noisy data. During optimization, the input point set $Q$ is generated from the raw point cloud through a probability distribution $\mathcal{D}$, which is built based on the neighborhood of the query points. Therefore, the density variation will affect the input data, and the generated points in sparse areas may be far away from the surface, increasing the difficulty of optimization in the point moving operation. We use the same parameters in all categories (clean, noise, and density variation) of a dataset for a fair comparison with other methods. A solution may be to choose different distributions for clean, noisy and uneven point clouds of a dataset, respectively.

### 7.4 How to obtain the correct normal orientation

The normal orientation is achieved by the gradient of the learned implicit function, and our proposed neural gradient function learns an implicit global surface representation from data. The implicit representation approaches, such as signed distance fields (SDF), represent the surface as zero level set of an implicit function $f$, *i.e.*, $f(x) = 0$. Therefore, we can train a neural network to regress singed distances, where SDF < 0 for inside, SDF > 0 for outside, and SDF = 0 for surface, so that the SDF increases from inside to outside of the surface. Then, the gradient vector field of the SDF is obtained, and the gradient on the iso-surface should have a uniform orientation.

In our method, the optimization is formulated as an iterative moving operation of points to the target surface. According to Neural-Pull [17], the gradient indicates the direction in 3D space in which the signed distance from the surface increases the fastest, so moving a point along or against (decided by SDF) the gradient will find its nearest path to the surface. We can obtain the gradient at each point using the learned SDF, and the gradient is perpendicular to the surface and points to inside or outside based on the initialization of the network.

### 7.5 Clarification of some figures

The output signs in Figure 3 and Figure 8 are not ignored, and the estimated normals are oriented and have consistent orientations. We know that implicit functions can reconstruct artifact surfaces from point clouds with open surface structures, but we do not care about the entire reconstructed surface, and only focus on the regions of existing points on the surface where the SDF can be correctly defined and its gradient has a consistent orientation. For a region without points, whose SDF is uncertain and the zero iso-surface is indeterminate, we do not use the SDF of this region to solve for the gradient. The example in Figure 3 is a part of a full shape with a closed surface, its inside/outside is defined, and we use a section of it for visualization. The example in Figure 8 is a point cloud of the KITTI dataset, the implicit function will learn a closed surface from it, and the points on the surface have a consistent gradient. We only use the SDF at points to solve for gradients as the normals.

### 7.6 Discussion of contributions and improvements to previous works

As we can see from Table 2 of the paper, existing supervised methods can achieve high-precision unoriented normal. However, from Table 1 of the paper, we can observe that higher-precision unoriented normal do not result in more accurate oriented normal using a normal orientation algorithm based on propagation strategies, such as PCA+MST *vs*. AdaFit+MST and PCA+SNO *vs*. HSurf-Net+SNO. This means that even if we develop better unoriented normal estimation algorithms, utilizing existing normal orientation algorithms will not necessarily lead to better orientation results. In brief, the bottleneck of oriented normal estimation is correctly determining the orientation. Moreover, the supervised methods require expensive ground truth as supervision and have been extensively studied, while unsupervised learning of normals is still an unexplored field. Based on the above two observations, we focus on how to use an unsupervised manner to directly learn oriented normals with higher orientation accuracy, instead of learning unoriented normals.

In this work, we introduce a neural gradient function, consisting of the multi-step moving strategy along the gradient in Eq.(2), the gradient uniformity of multi-scale neighborhood size in Eq.(7), and the gradient consistency of multi-step moving in Eq.(8). From the ablation studies, we can see that the performance of the algorithm drops a lot if we do not use these strategies, especially the losses in Eq.(7) and Eq.(8). In Fig.6 of the paper and Fig.1 of the supplementary material, thanks to the losses in Eq.(2) and Eq.(7), our method is more robust against noise compared to surface reconstruction methods.

Existing learning-based normal estimation methods, such as PCPNet, require ground truth normals as supervision for training, and there is still a lot of room for improvement in performance, especially in some challenging cases. Although PCPNet can predict point normal in a forward process with pre-trained parameters, it does not generalize well to unseen cases. We resolve this issue using a per-shape overfitting strategy, which significantly improves the performance in unseen cases. Our method directly learns normals from raw data without using ground truth and achieves better performance for various inputs. In addition, our method can also provide better surface reconstruction results. We do not intend to replace methods, such as PCPNet, that can directly use pre-trained models on different data, but rather as a new exploration that can provide another option for the 3D computer vision community to easily obtain more accurate normals and surfaces from point clouds.

### 7.7 The connection between surface reconstruction methods and normal estimation

In recent years, researchers have paid more attention to the global consistency of normal orientations, such as iterative Poisson Surface Reconstruction (iPSR) [9], Parametric Gauss Reconstruction (PGR) [15] and Stochastic Poisson Surface Reconstruction (SPSR) [23]. In addition to traditional approaches, deep neural networks have been applied to gather information for orientation and reconstruction. Some works learn the implicit function directly from the input point cloud and

9

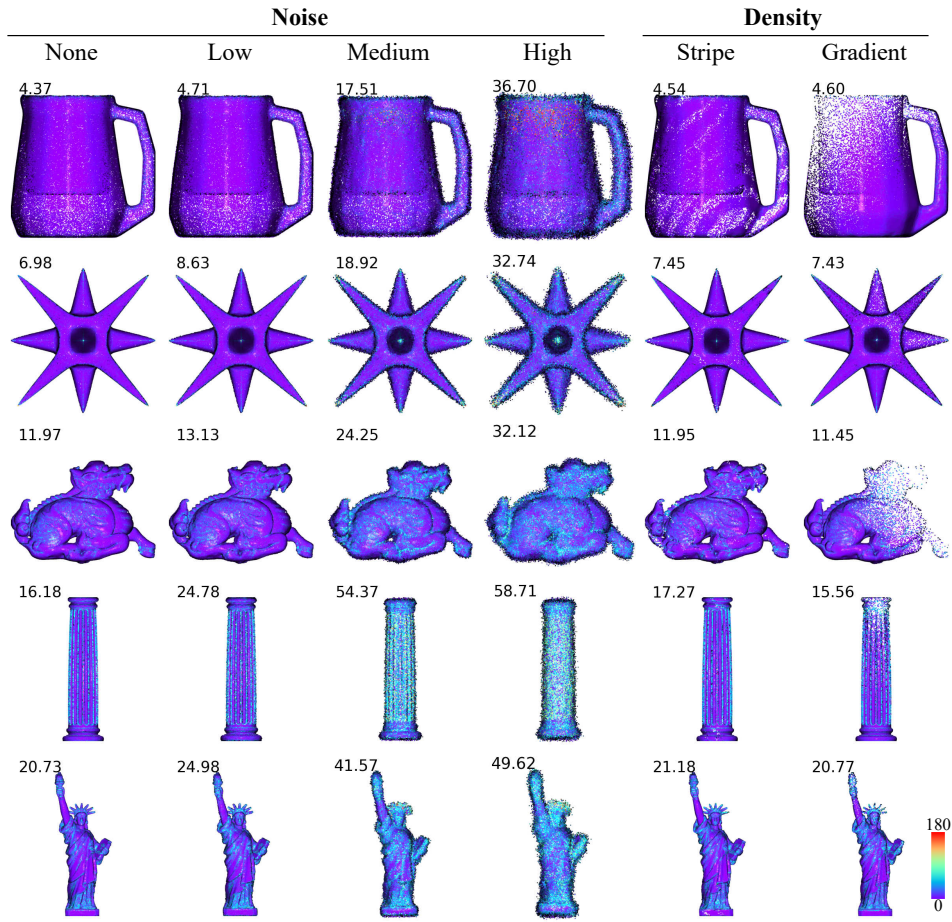

Figure 10: Error visualization of oriented normals estimated by our method on several shapes of the PCPNet dataset [7]. We map the normal errors to a heatmap ranging from $0°$ (purple) to $180°$ (red) for visualization. The average RMSE values are reported for each point cloud.

eliminate the need for training data, such as SAL [1], Neural-Pull [17], IGR [6] and SAP [21]. Although these surface reconstruction methods do not aim to estimate point cloud normals, they will add normals (or gradients) to the constraint conditions during the surface optimization process to assist in surface reconstruction. The experimental results of these methods also validate the benefits of using normals in surface reconstruction.

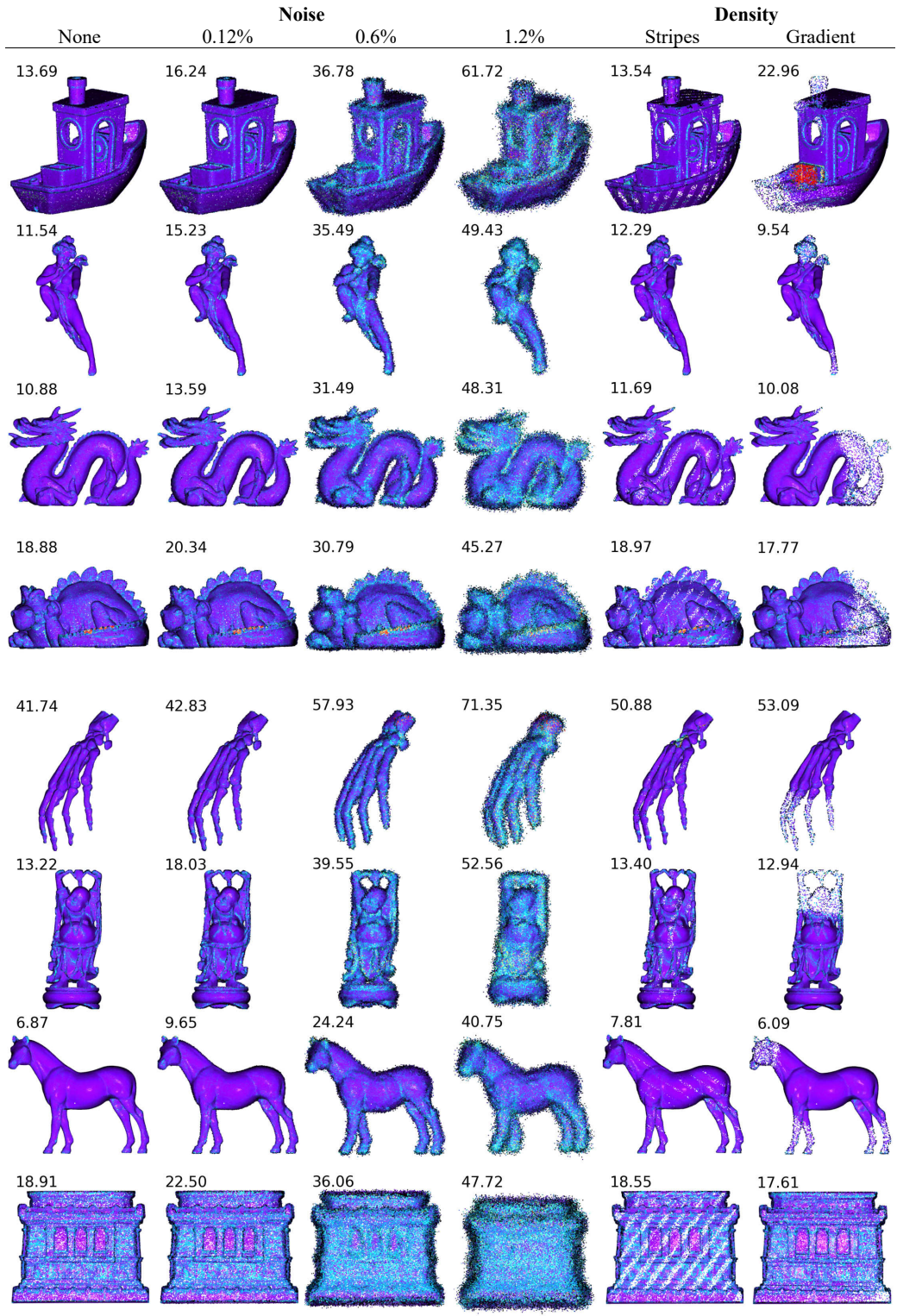

Figure 11: Error visualization of oriented normals estimated by our method on the FamousShape dataset [12]. We map the normal errors to a heatmap ranging from $0°$ (purple) to $180°$ (red) for visualization. The average RMSE values are reported for each point cloud.

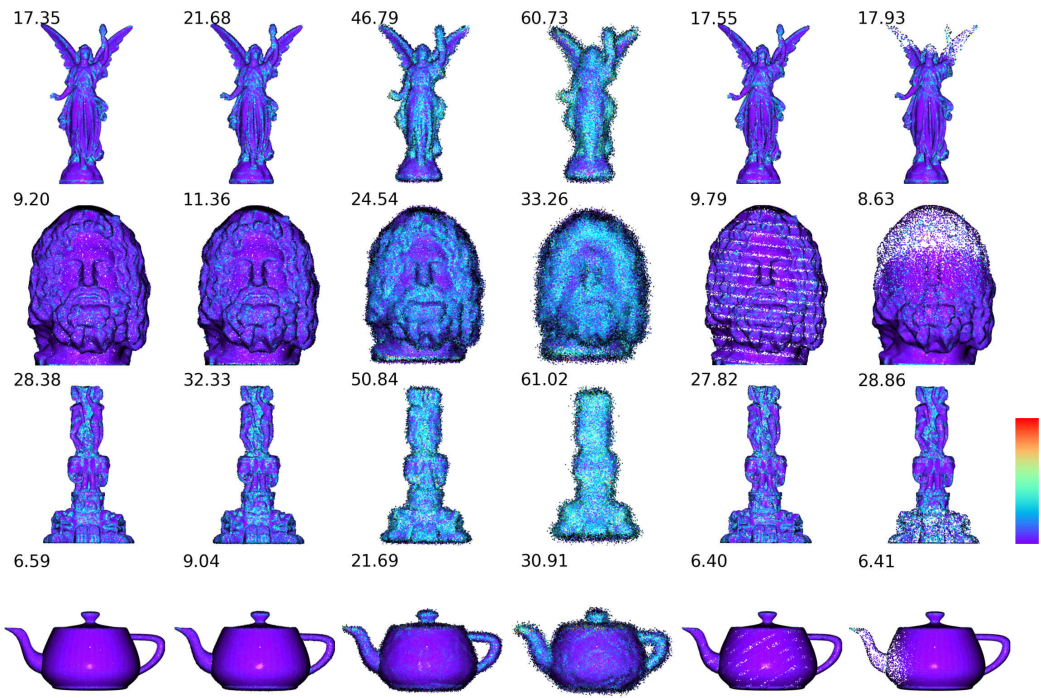

Figure 12: (Same as the previous figure.) Error visualization of oriented normals estimated by our method on the FamousShape dataset [12]. We map the normal errors to a heatmap ranging from $0°$ (purple) to $180°$ (red) for visualization. The average RMSE values are reported for each point cloud.

13



\end{document}